\documentclass[aos,preprint,reqno]{imsart}

\setattribute{journal}{name}{}

\usepackage[utf8]{inputenc} 
\usepackage[T1]{fontenc}    
\usepackage[margin=1.1in]{geometry}
\usepackage{hyperref}       
\usepackage{url}            
\usepackage{booktabs}       
\usepackage{amsfonts}       
\usepackage{nicefrac}       
\usepackage{microtype}      

\usepackage{natbib}
\usepackage{amsmath, amssymb, amsthm, bbm, dsfont, mathrsfs}
\usepackage{amsthm}
\usepackage{subfiles, subcaption}
\usepackage{color, verbatim, graphicx}
\usepackage{algorithm, algpseudocode}

\usepackage{cleveref}
\usepackage{mathtools}
\usepackage{commath}
\usepackage{stmaryrd}
\usepackage[symbol]{footmisc}
\usepackage{subfiles}

\let\norm\undefined 
\DeclarePairedDelimiter\norm{\lVert}{\rVert}

\newcommand{\E}{\mathbb{E}}
\newcommand{\nats}{\mathbb{N}}

\newcommand{\reals}{\mathbb{R}}

\newcommand{\GP}{\mathrm{GP}}

\usepackage{xspace}
\usepackage{xr}
\newcommand{\X}{\mathcal{X}}
\newcommand{\N}{\mathcal{N}}
\newcommand{\given}{\,|\,}

\begin{document}

\begin{frontmatter}

\title{Multi-fidelity Monte Carlo: \\ a pseudo-marginal approach}
\runtitle{Multi-fidelity Monte Carlo: a pseudo-marginal approach}

\begin{aug}
\author{\fnms{Diana} \snm{Cai}\thanksref{paddr}}
\and
\author{\fnms{Ryan P.} \snm{Adams}\thanksref{paddr}}

\affiliation{Princeton University\thanksmark{paddr}}

\address[paddr]{
    Department of Computer Science \\
    Princeton University \\
    Princeton, NJ, USA 08544\\
    Emails: \href{mailto:dcai@cs.princeton.edu}{dcai@cs.princeton.edu},
    \href{mailto:rpa@princeton.edu}{rpa@princeton.edu}
}

\runauthor{D.\ Cai and R.P.\ Adams}
\end{aug}


\begin{abstract}
    Markov chain Monte Carlo (MCMC) is an established approach for uncertainty quantification and propagation in scientific applications.  A key challenge in applying MCMC to scientific domains is computation: the target density of interest is often a function of expensive computations, such as a high-fidelity physical simulation, an intractable integral, or a slowly-converging iterative algorithm.  Thus, using an MCMC algorithms with an expensive target density becomes impractical, as these expensive computations need  to be evaluated at each iteration of the algorithm.  In practice, these computations often approximated via a cheaper, low-fidelity computation, leading to bias in the resulting target density.  Multi-fidelity MCMC algorithms combine models of varying fidelities in order to obtain an approximate target density with lower computational cost.  In this paper, we describe a class of asymptotically exact multi-fidelity MCMC algorithms for the setting where a sequence of models of increasing fidelity can be computed that approximates the expensive target density of interest.  We take a pseudo-marginal MCMC approach for multi-fidelity inference that utilizes a cheaper, randomized-fidelity unbiased estimator of the target fidelity constructed via  random truncation of a telescoping series of the low-fidelity sequence of models.  Finally, we discuss and evaluate the proposed multi-fidelity MCMC approach on several applications, including log-Gaussian Cox process modeling, Bayesian ODE system identification, PDE-constrained optimization, and Gaussian process regression parameter inference.

\end{abstract}

\end{frontmatter}

\maketitle


\section{Introduction}

Simulation and computational modeling play a key role in science, engineering, economics, and many other areas.
When these models are high-quality and accurate, they are important for scientific discovery, design, and data-driven decision making.
However, the ability to accurately model complex physical phenomena often comes with a significant cost---many models involve expensive computations that then need to be evaluated repeatedly in, for instance, a sampling or optimization algorithm.
Examples of model classes with expensive computations include intractable integrals or sums, expensive quantum simulations \citep{troyer2005computational},
expensive numerical simulations arising from partial differential equations (PDEs) \citep{raissi2017inferring} and large systems of ordinary equations (ODEs).

In many situations, one has the ability to trade off computational cost against \emph{fidelity} or accuracy in the result.
Such a tradeoff might arise from the choice of discretization or the number of basis functions when solving a PDE, or the number of quadrature points when estimating an integral.
It is often possible to leverage lower-fidelity models to help accelerate high-quality
solutions, e.g., by using multigrid methods \citep{hackbusch2013multi} for spatial discretizations.
More generally, \emph{multi-fidelity} methods combine multiple models of varying cost and fidelity to accelerate computational algorithms and have been
applied to solving inverse problems \citep{higdon2002bayesian,cui2015data,raissi2017inferring},
trust region optimization \citep{alexandrov1998trust,arian2000trust,fahl2003reduced,robinson2008surrogate,march2012constrained},
Bayesian optimization \citep{jones1998efficient,gramacy2009adaptive,song2019general,wu2020practical,li2020multi,brevault2020overview},
Bayesian quadrature \citep{gessner2020active,xi2018bayesian},
and
sequential learning \citep{gundersen2021active,palizhati2022agents}.

One critically important tool for scientific and engineering computation is Markov chain Monte Carlo (MCMC), which is widely used for uncertainty quantification, optimization, and integration.
MCMC methods are recipes for constructing a Markov chain with some desired target distribution as the limiting distribution.
Pseudo-random numbers are used to simulate transitions of the Markov chain in order to produce samples from the target distribution.
However, MCMC often becomes impractical for high-fidelity models, where a single step of the Markov chain may, for instance, involve a numerical simulation that takes hours or days to complete.
Multi-fidelity methods for MCMC focus on constructing Markov chain transition operators that are sometimes able to use inexpensive low-fidelity evaluations instead of expensive high-fidelity evaluations.
The goal is to increase the effective number of samples generated by the algorithm, given a constrained computational budget.
A large focus of the multi-fidelity MCMC literature is on two-stage Metropolis-Hastings (M-H)
methods \citep{christen2005markov,efendiev2006preconditioning}, which use a single low-fidelity model for early rejection of a proposed sample, thereby often short-circuiting the evaluation of the expensive, high-fidelity model.

However, there are several limitations of two-stage multi-fidelity Monte Carlo.
First, in many applications, a \emph{hierarchy} of cheaper, low-fidelity models is available; for instance, in the case of integration, $k$-point quadrature estimates form a hierarchy of low-fidelity models, and in the case of a PDE, varying the discretization.
Thus, the two-stage approach does not fully utilize the availability of a hierarchy of fidelities and may be more suitable for settings where the high- and low-fidelity models are not hierarchically related, e.g., semi-empirical methods vs.\ Hartree-Fock in computational chemistry.
In addition, in such applications,
there is often a limiting model of interest, such as a continuous function that the low-fidelity discretizations approximate.
Two-stage MCMC does not asymptotically sample from this limiting target density and will at best sample from an approximation of the biased, high-fidelity posterior.
Finally, the two-stage method is unnatural to generalize to more sophisticated
MCMC algorithms such as slice sampling and Hamiltonian Monte Carlo (HMC).
%

\begin{figure}[t]
    \centering
    \begin{subfigure}[b]{0.50\linewidth}
        \centering
        \includegraphics[scale=0.54]{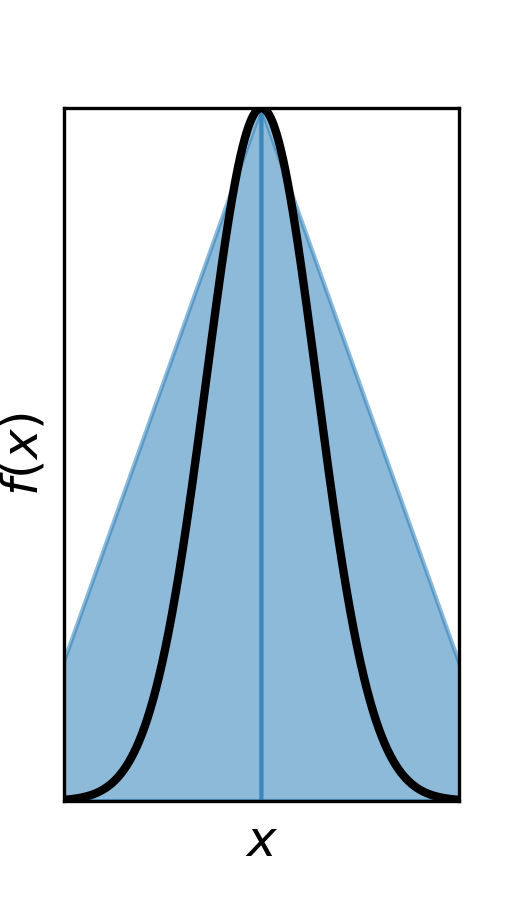}
        \includegraphics[scale=0.54]{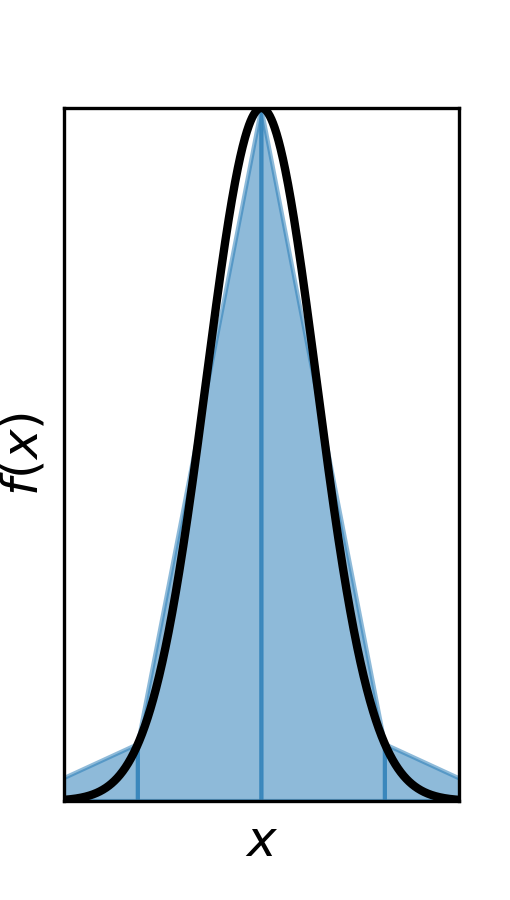}
        \includegraphics[scale=0.54]{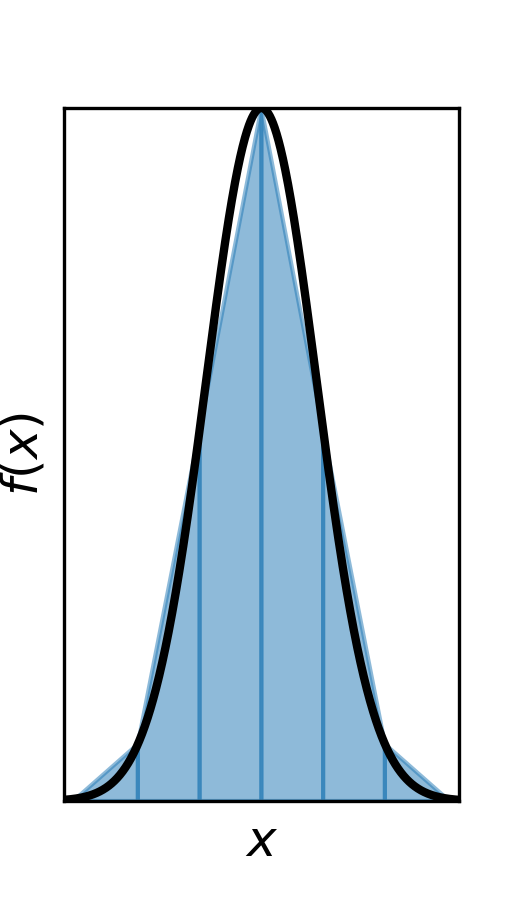}
        \vspace{8pt}
        \caption{Trapezoid rule with $2k$ trapezoids ($k=1,2,3$)}
    \end{subfigure}
    \begin{subfigure}[b]{0.49\linewidth}
        \centering
        \includegraphics[scale=0.62]{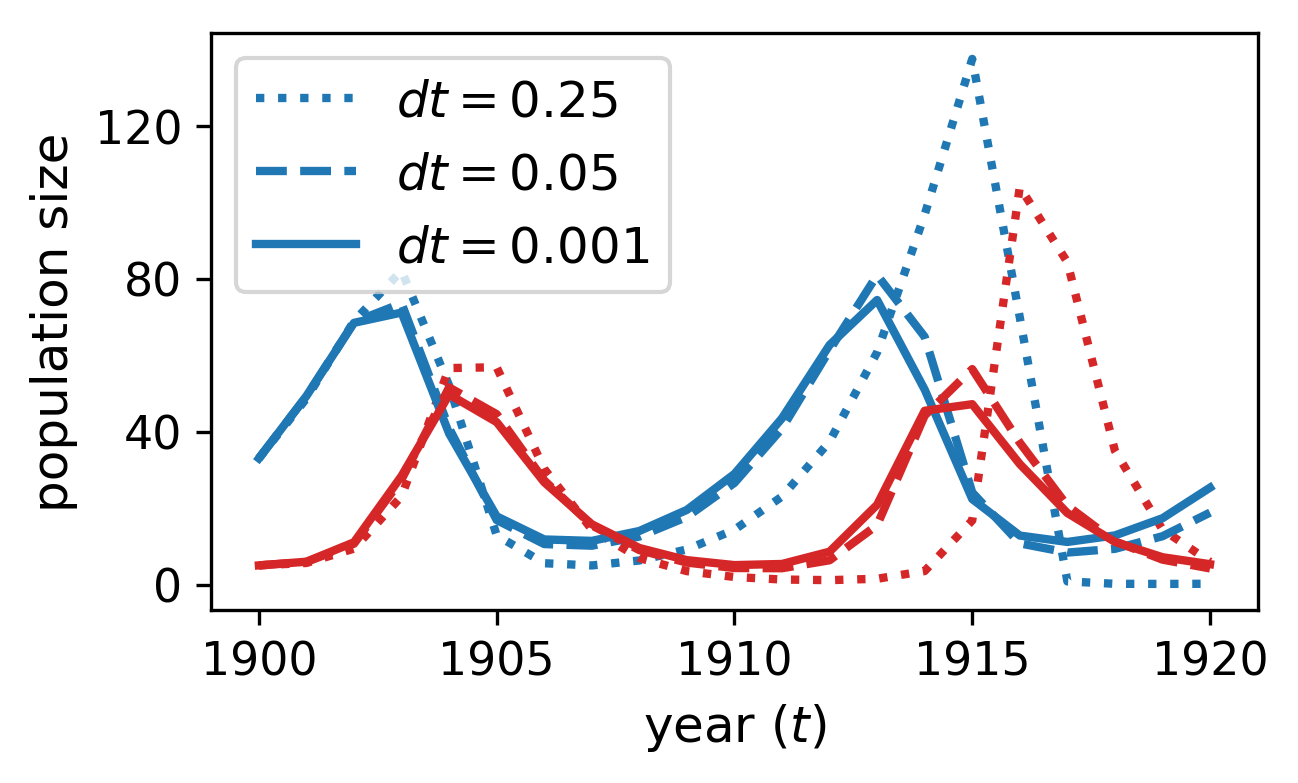}
        \caption{Lokta-Volterra ODE solutions, $dt=1/k$}
    \end{subfigure}
    \caption{\small Examples of low-fidelity sequences of models.
    \textbf{(a)} Sequence trapezoid quadrature estimates $I_k$,
    where $I_k$ is the trapzoid rule with $2k$ trapezoids.
    \textbf{(b)} Lokta-Volterra ODE solutions for prey $u(t)$ (blue) and predator $v(t)$
    (red) using Euler's method with step size $dt$.
    }
    \label{fig:fidelities}
    \vspace{-12pt}
\end{figure}

We propose a class of multi-fidelity MCMC methods designed for applications with a hierarchy of low-fidelity models available.
More specifically, we assume access to a sequence of low-fidelity models that
converge to a ``perfect-fidelity'' model in the limit.
Within an MCMC algorithm, we can approximate the perfect-fidelity target density with an unbiased estimator constructed from a randomized truncation of the infinite telescoping series of low-fidelity target densities.
This class of multi-fidelity MCMC is an example of a pseudo-marginal MCMC
(PM-MCMC) algorithm---the unbiased estimator
essentially guarantees that the algorithm is asymptotically exact in that the
limiting distribution recovers the perfect-fidelity target distribution as its marginal distribution.
Our approach introduces the fidelity of a model as an auxiliary random variable
that is evolved separately from the target variable according to its own
conditional target distribution; this technique can be used in conjunction with
any suitable MCMC update that leaves the conditional update for the target
variable of interest invariant, such as M-H, slice sampling, elliptical slice sampling, or Hamiltonian Monte Carlo.
We apply the pseudo-marginal multi-fidelity MCMC approach to several problems, including
log-Gaussian Cox process modeling, Bayesian ODE system identification, PDE-constrained optimization, and Gaussian process parameter inference.

\subsection{Related work}
Multi-fidelity MCMC methods are commonly applied in a two-stage procedure,
where the goal is to reduce the computational cost of using a single expensive high-fidelity model
by using a cheap low-fidelity model as a low-pass filter
for a delayed acceptance/rejection algorithm
\citep{christen2005markov,efendiev2006preconditioning,cui2015data};
see \citet{peherstorfer2018survey} for a survey.
\citet{higdon2002bayesian} propose coupling a high-fidelity Markov chain with a low-fidelity
Markov chain via a product chain.
In constrast, our approach aims to sample from a ``perfect-fidelity'' target density while reducing
computational cost;
two-stage MCMC algorithms result in biased estimates with respect to
this target density.
A related class of methods is multilevel Monte Carlo
\citep{giles2008multilevel,giles2013multilevel,dodwell2015hierarchical,warne2021multifidelity}, which uses a hierarchy of multi-fidelity models
for Monte Carlo estimation by expressing the expectation of a high-fidelity model
as a telescoping sum of low-fidelity models.
\citet{dodwell2015hierarchical} use the M-H algorithm to form the multilevel Monte
Carlo estimates,
simulating from a separate Markov chain for each level of the telescoping sum.
In practice multilevel Monte carlo requires choosing a finite
number of fidelities, inducing bias in the estimator with respect to the (limiting) perfect-fidelity
model.
In contrast, our method uses a randomized fidelity within a single Markov chain
with the perfect-fidelity model as the target.

Our approach applies pseudo-marginal MCMC to multi-fidelity problems.
There is a rich literature developing pseudo-marginal MCMC methods \citep{beaumont2003estimation,andrieu2009pseudo}
for so-called ``doubly-intractable'' likelihoods,
which are likelihoods that are  intractable to evaluate.
Several approaches in the pseudo-marginal MCMC literature are particular relevant to our
work.
The first are the PM-MCMC methods introduced by \citet{lyne2015russian}, which
describes a class of pseudo-marginal M-H methods that use Russian roulette estimators to obtain
unbiased estimators of the likelihood.
However, this method samples the variable of interest jointly with the auxillary randomness,
which often leads to sticking.

Alternatively, several methods have considered sampling the randomness separately.
The idea of clamping random numbers is explored in depth by \citet{leeholmes2010} and
\citet{murray2016pseudo}; the latter applies to this pseudo-marginal slice sampling.
In particular, our approach applies these ideas to the specific setting of multi-fidelity models,
where the random fidelity is treated as an auxillary variable.
Finally, while our approach applies to doubly-intractable problems, we are also motivated
by a larger class of multi-fidelity problems studied in the computational sciences that may not
even be inference problems, such as quantum simulations and PDE-constrained
optimization.


\section{Multi-fidelity MCMC}
\label{sec:mfmcmc}

Monte Carlo methods approximate integrals and sums that can be expressed an expectation:
\begin{align}
\label{eq:functional}
    \E_{\pi}(h(\theta)) =
\int h(\theta)\, \pi(\theta)\,d\theta &\approx \frac{1}{T}\sum_{t=1}^T
    h(\theta^{(t)}), \qquad \text{where} \quad \theta^{(t)}\sim \pi,
\end{align}
and where $\pi: \Theta \rightarrow \reals_+$ is the \emph{target density} that
may only be known up to a constant,
$h(\theta)$~is a function of interest,
and $\{\theta^{(t)}\}_{t=1}^T$ are samples from~$\pi$.
Markov chain Monte Carlo methods are then used to generate samples
$\theta^{(t)}$ from~$\pi$ by simulating from a Markov chain
with target $\pi$.

In many settings, pointwise evaluations of the target function $\pi(\theta)$ are
expensive or even intractable;
from here on we will assume that the goal is to compute statistics
of a quantity of interest $h(\theta)$
with respect to
a \emph{perfect-fidelity} target density $\pi_\infty(\theta)$.
In practice, the estimate in \Cref{eq:functional}
is instead estimated using a cheaper, low-fidelity density
$\pi_k(\theta)$, where $k \in \nats := \{1,2,\ldots\}$.
In particular, we consider settings where
there is a \emph{sequence} of low-fidelity densities available that converge to the target, i.e.,
$\pi_k(\theta) \stackrel{k \rightarrow\infty}{\longrightarrow}
\pi_\infty(\theta)$.
We assume that as $k$ increases, the model becomes higher in fidelity (with respect
to $\pi_\infty$)
but more costly to evaluate, increasing in expense super-linearly with $k$.

For instance, $\pi_\infty$ could represent a target density that depends on
an intractable integral,
the solution of a PDE, the solution of a large system of ODEs,
the solution of a large system of linear equations,
or the minimizer of a function.
Thus, a typical evaluation of $\pi_\infty$ requires an approximation at a fidelity $k$ with
a tolerable level of bias for a given computational budget.
Here increasing~$k$ could correspond to finer discretizations of differential equations,
increasing numbers of quadrature points, or performing a larger number of iterations
in a linear solver or optimization routine.

In the multi-fidelity setting, the goal is to combine several models
of varying fidelity within an MCMC algorithm to reduce the computational cost of estimating
\Cref{eq:functional}.
In this paper, we describe a class of MCMC algorithms that leverages
the sequence of low-fidelity models $\pi_k$.
Our strategy for multi-fidelity MCMC (MF-MCMC) will be to construct an unbiased estimator
of~$\pi_\infty(\theta)$ using random choices of the fidelity~$K$ and then to include~$K$
in the Markov chain as an auxiliary variable.
By carefully constructing such a Markov chain, it will be possible to asymptotically estimate the
functional in~\Cref{eq:functional} as though the samples were taken from the perfect-fidelity
model; each step of the Markov chain will nevertheless only require a finite amount of computation.
Finally, our approach allows us to essentially plug in any valid MCMC algorithm, and
we apply this strategy to develop multi-fidelity variants of a number of MCMC algorithms,
such as M-H and slice sampling.

\subsection{Pseudo-marginal MCMC for the multi-fidelity setting}
\label{ssec:mainmethod}

Pseudo-marginal MCMC \citep{beaumont2003estimation,andrieu2009pseudo} is a class
of auxillary-variable MCMC algorithms
that replaces the target density~$\pi(\theta)$ with an estimator $\hat{\pi}(\theta)$ that is a
function of a random variable.
If the estimator is nonnegative and unbiased, i.e.,
for all $\theta \in \Theta$,
    $\hat{\pi}(\theta) \geq 0$ and
$\mathbb{E}[\hat{\pi}(\theta)] = \pi(\theta)$,
then MCMC transitions that use the estimator still have~$\pi(\theta)$ as their
invariant distribution.
This property is sometimes referred to as ``exact-approximate'' MCMC as the transitions are
approximate but the limiting distribution is exact.
Estimators can be constructed from a variety of methods, including particle filtering
\citep{andrieu2009pseudo}; our approach will
use randomized series truncations, which has been consider in pseudo-marginal
MCMC methods such as
\citet{lyne2015russian}, \citet{georgoulas2017unbiased}
and \citet{biron2021pseudo}.

We now apply the pseudo-marginal
approach to the multi-fidelity setting.
Here the target density estimator arises from a random choice of the fidelity~$K \in \nats$
that is governed by a distribution $\mu$ on $\nats$.
We denote the estimator using
$\hat{\pi}_K(\theta)$ to make the dependence on the random fidelity $K$
explicit.
The estimator is constructed such that it is
unbiased with respect to $\mu$, i.e.,
\begin{align}
\label{eq-unbiased}
\sum_{k=1}^\infty \mu(k) \hat{\pi}_k(\theta) &= \pi_\infty(\theta)\,.
\end{align}

The distribution $\mu$ is also constructed by the user:
ideally, the estimator~$\hat{\pi}_K(\theta)$ will prefer smaller values of~$K$
while having
sufficiently low variance as to allow the Markov chain to mix effectively.
Thus the simulations can
be run at inexpensive low-fidelities, while the estimates will be as though
the perfect-fidelity model were being used.

The standard pseudo-marginal MCMC approach is to construct a Markov chain
that has the following joint density as its stationary distribution:
\begin{align}
    \label{eq-joint-density}
    \pi(\theta, K) =
    \mu(K) \hat{\pi}_K(\theta).
\end{align}
Observe that while \Cref{eq-joint-density}
does not depend on the perfect-fidelity target density
$\pi_\infty$,
it returns the desired marginal $\pi_\infty$ via
\Cref{eq-unbiased}.
As a concrete example, a pseudo-marginal M-H algorithm generates
a new state $\theta'$ and fidelity~$K'$ jointly using~$q(\theta';\theta)$ as the proposal for~$\theta'$,~$q(K'; K) = \mu(K')$
as the proposal distribution for the fidelity,
and accepts/rejects the state according to
\begin{align}
    \label{eq-MH}
    a =
    \frac{\pi(\theta', K') q(\theta;\theta')q(K;K')}{\pi(\theta, K) q(\theta';\theta)q(K';K)}
    =
    \frac{\hat\pi_{K'}(\theta') q(\theta;\theta')}{\hat\pi_K(\theta) q(\theta';\theta)},
\end{align}
where the equality holds since the distribution terms for $K$ and $K'$ cancel.
Note that the right-hand side of \Cref{eq-MH} is the standard M-H
ratio but that the target density $\pi$ is replaced with the estimator $\hat\pi_K$.

However, standard pseudo-marginal MCMC using joint proposals of the state and
fidelity can ``get stuck'' when
the estimator is noisy and fail to accept new states.
Thus, we apply the approach in \citet{murray2016pseudo} that augments the Markov
chain to include the randomness of the estimator via a separate update;
here the randomness of the estimator arises from the fidelity
$K$.
Concretely, we construct a Markov chain that simulates from \Cref{eq-joint-density}
by alternating sampling between the conditional target densities
$\pi(K | \theta)$ and $\pi(\theta | K)$ (steps 5 and 6 of
\Cref{alg:MF-MCMC-sign}, respectively).
We refer to this strategy as \emph{multi-fidelity MCMC} (MF-MCMC), since
by conditioning on $K=k$,
the update for the state $\theta$
becomes a standard deterministic update applied to a low-fidelity model
$\hat\pi_k(\theta)$,
and any appropriate MCMC update can be used here, making it straightforward to
use complex MCMC methods, such as slice sampling and HMC.
Similarly, any suitable MCMC update for the fidelity $K$ can be used using the
conditional target~$\pi(K | \theta)$.

Many techniques can be used to construct an unbiased estimator of $\pi_\infty$
with randomness $K$; we describe a general approach in the next section.
However, it is generally difficult to guarantee the estimator is nonnegative,
as required by pseudo-marginal MCMC.
One technique considered by \citet{lin2000noisy} and \citet{lyne2015russian} is
to instead sample from the target distribution induced by the absolute value of the
estimator and applying a sign-correction to the final Monte Carlo
estimate in \Cref{eq:functional}, an approach borrowed from the quantum Monte
Carlo literature where it is necessary for modeling fermionic particles.
This approach has been applied to the M-H algorithm, but we note that this
general approach can be applied much more broadly, as we do in this work.

In problems where the estimator may be negative,
we sample from the conditional target distributions using the absolute value of
the estimator $|\hat{\pi}_K(\theta)|$, and we denote these conditionals with
$\tilde \pi(K \given \theta) \propto \mu(K) |\hat{\pi}_K(\theta)|$
and $\tilde\pi(\theta | K=k) \propto |\hat{\pi}_k(\theta)|$.
The estimate in \Cref{eq:functional} is then corrected using the signs
$\sigma(\theta, k)$ of evaluations of $\hat\pi_k(\theta)$,
\begin{align}
    \label{eq:signcorrection}
    \!\int\! h(\theta)\, \pi(\theta) \,d\theta
    \approx
    \frac{\sum_{t=1}^T   h(\theta^{(t)})  \sigma(\theta^{(t)}, K^{(t)})
         }
    {\sum_{t=1}^T \!  \sigma(\theta^{(t)}, K^{(t)})
         }
         =: \hat{I}_T,
\end{align}
where $\{(\theta^{(t)},K^{(t)})\}_{t=1}^T$ are the sampled values
from the joint distribution~$\tilde \pi(\theta, K) \propto |\hat{\pi}_K(\theta)| \mu(K)$.

Importantly, the sign-corrected estimate still asymptotically leads to
the desired estimate of the functional of interest.
Let $\sigma(\theta, k)$ denote the sign of the estimator such that
$\hat{\pi}_k(\theta) = \sigma(\theta, k) |\hat{\pi}_k(\theta)|$.
The estimator $\hat{I}_T$ in \Cref{eq:signcorrection}
is formed using a
Monte Carlo estimate of the functional after expanding it into its joint
distribution, i.e.,
\begin{align}
    \label{eq-sign-correction-deriv}
    \int h(\theta) \pi_\infty(\theta) d\theta
    &=
    \int \sum_{k = 1}^\infty  h(\theta)  \hat{\pi}_k(\theta) \mu(k) d\theta
    =
    \frac{\int \sum_{k = 1}^\infty  h(\theta)  \sigma(\theta, k) \tilde\pi(\theta, k)  d\theta }
    {\int \sum_{k = 1}^\infty \sigma(\theta,k) \tilde\pi(\theta, k)  d\theta
    }.
\end{align}

The full multi-fidelity MCMC algorithm with sign correction
summarized in
\Cref{alg:MF-MCMC-sign}.
We note that while the Markov chain no longer converges to a target with the
marginal $\pi_\infty$,
the final estimate after sign-correction---which is the downstream
goal of interest---converges to the quantity of interest  due to \Cref{eq-sign-correction-deriv}.
While this may seem limiting if one is interested in the posterior itself,
useful unbiased posterior summaries may be still be obtained via the functional,
such as the posterior mean, variance, quantiles, and histograms that may be used to
visualize marginal distributions.

\section{Unbiased low-fidelity estimators via randomized truncations}
\label{ssec:estimator}

In this section, we discuss how to construct an unbiased estimator of $\pi_\infty(\theta)$, given a sequence of low-fidelity likelihoods with the property $\pi_k(\theta) \rightarrow \pi_\infty(\theta)$ as $k \rightarrow \infty$.
This estimator has the property that it requires a finite amount of computation with probability one, and it also has a tunable amount of expected computation per estimate, i.e., it uses low-fidelity density evaluations to estimate the perfect-fidelity target density.
The central idea of this estimator has been used for decades, going back to John von Neumann and Stanislaw Ulam.
More recently it has found use in applications of inference and optimization in related work
such as \citet{glynn2014exact}, \citet{lyne2015russian}, \citet{beatson2019efficient},
and \citet{jacob2020unbiased}.

First note that we can express the perfect-fidelity model as a telescoping
sum of low-fidelity models: let $\pi_0(\theta) = 0$ and write
\begin{align}
    \label{eq-series}
    \pi_\infty(\theta) = \sum_{k=1}^\infty \pi_k(\theta) - \pi_{k-1}(\theta).
\end{align}
The estimator $\hat{\pi}_K$ is then constructed by
taking a random truncation $K \sim \mu$ of the infinite telescoping series.
The sampled terms in the sum are then reweighted
to ensure the estimator remains unbiased:
\begin{align}
    \label{eq-truncated-series}
    \hat{\pi}_K(\theta) =
    \sum_{k=1}^K w_{k,K} (\pi_k(\theta) - \pi_{k-1}(\theta))\,.
\end{align}
Two approaches are commonly used to ensure that
the resulting estimator is unbiased:
weighted single-term estimators and Russian roulette estimators.
The single-term estimator \citep{lyne2015russian} is constructed by importance sampling
a term from the series in \Cref{eq-series}: the truncation level is drawn as $K \sim \mu$, and the $K$th term
is used to form the estimate
\begin{align}
\hat{\pi}_K(\theta) =  \mu(K)^{-1} (\pi_K(\theta) - \pi_{K-1}(\theta))\,.
\end{align}
Thus, the weight in \Cref{eq-truncated-series} is ${W_{k,K} = \mu(K)^{-1}
\mathbbm{1}(K=k)}$.
In the Russian roulette estimator, the remaining terms in the estimator are reweighted
by their survival probabilities, i.e.,~${W_{k,K} = (1-\sum_{k'=1}^{k-1} \mu(k'))^{-1} \mathbbm{1}(K \geq k)}$.
The distribution~$\mu$ controls the number of terms in the estimator,
and
a good proposal distribution should match the tails of the sequence of
low-fidelity densities
\citep{lyne2015russian,beatson2019efficient,potapczynski2021bias}.

The ability to use cheaper models is a key feature of multi-fidelity inference,
and the low-fidelity estimator provides a means to reduce the computational cost of multi-fidelity
Monte Carlo.
However, these estimators are an example of a class of methods that
explores a compute-variance tradeoff:
computationally cheaper estimates leads to high variability.
The resulting increase in variance
slows down the convergence of the MCMC procedure and could
lead to an overall less efficient method due to a reduced effective sample size.

\begin{algorithm}[t!]
\caption{Multi-fidelity Monte Carlo with sign-correction}
\label{alg:MF-MCMC-sign}
\begin{algorithmic}[1]
\State \textbf{Input}: Initial state $\theta$ and fidelity $K$, 
truncation distribution $\mu$
    \For{$t=1,\ldots,T$}
        \State
        Given current $K$ and $\theta$, form estimator
        \[\hat{\pi}_K(\theta) =
        \sum_{k=1}^K w_{k,K} (\pi_k(\theta) - \pi_{k-1}(\theta))\]
        \State Save sign $\sigma(\theta, K) = \text{sign}(\hat{\pi}_K(\theta))$
        \State Update fidelity $K$ leaving invariant the target conditional
            \[\tilde\pi(K | \theta) \propto \mu(K) |\hat{\pi}_K(\theta)|\]
        \State Update state $\theta$ leaving invariant the target conditional 
            \[\tilde \pi(\theta | K=k) \propto |\hat{\pi}_k(\theta)|\]
    \EndFor
\State \textbf{Output}:
    Samples $\{(\theta^{(t)}, K^{(t)})\}$ and
    estimate $\hat{I}_T =
    \left(\sum_{t=1}^T  \sigma^{(t)} h(\theta^{(t)})  \right) /
    \left({\sum_{t=1}^T \sigma^{(t)}}\right)$
\end{algorithmic}
\end{algorithm}

\section{Summary of the multi-fidelity MCMC recipe}
Here we summarize the recipe for constructing a multi-fidelity Markov chain Monte Carlo algorithm.

First, identify a sequence of increasing-fidelity target densities with the property that their limit is the desired ``perfect-fidelity'' density.
Low-fidelity densities should be cheap with the cost rapidly increasing within the sequence.
In the context of Bayesian inference, it may be appropriate to focus the
multi-fidelity aspects on the likelihood term and construct the target densities
via, e.g.,~$\pi_k(\theta; \mathcal{D}) \propto \pi_0(\theta) L_k(\theta\,;\,\mathcal{D})$,
where $\pi_0$ is the prior, $L_k$ is a low-fidelity likelihood, and
$\mathcal{D}$ is the set of observations.
This likelihood-based version is what we use in several of the experiments.

Next, introduce a truncation distribution~$\mu$ on $\mathbb{N}$.
This truncation distribution should be chosen to balance between expected cost and
variance of the resulting estimator; our overall goal is to mostly use cheap
low-fidelity densities, but high-variance estimators will presumably damage the
mixing time and/or the asymptotic variance.

Initialize the Markov chain with a reasonable choice for~$\theta$ and a draw
of~$K$ from the distribution~$\mu$.
Each step of the Markov chain simulation consists of an update to~$\theta$
given~$K$ and an update of~$K$ given~$\theta$.
The update of~$\theta$ given~$K$ can be performed using any standard MCMC algorithm,
e.g., M-H, slice sampling, or HMC, applied to the low-fidelity estimator.
It is important to use the absolute value of the estimator and keep
track of its sign.
The update of~$K$ given~$\theta$ is also flexible, but it is reasonable to
construct the update so that only a few~$K$ are considered in each step, as each of
those fidelities will need to be evaluated.
By default, we consider a simple random walk on the positive integers for our
experiments.
After running a sufficient number of steps of the Markov chain, use the sign
corrected-estimator in \Cref{eq:signcorrection} to compute the expectation of
the function~$h(\theta)$.


\section{Experiments}
\label{sec-experiments}

In all experiments, we use a random-walk M-H update to sample from the conditional
$K | \theta$, and truncation distribution $\mu(K) = \text{geometric}(K;
\gamma_0)$.
Additional experimental details are in \Cref{appendix:expts}.

\begin{figure}[t]
    \centering
    \begin{subfigure}[b]{0.22\linewidth}
        \centering
        \includegraphics[scale=0.51]{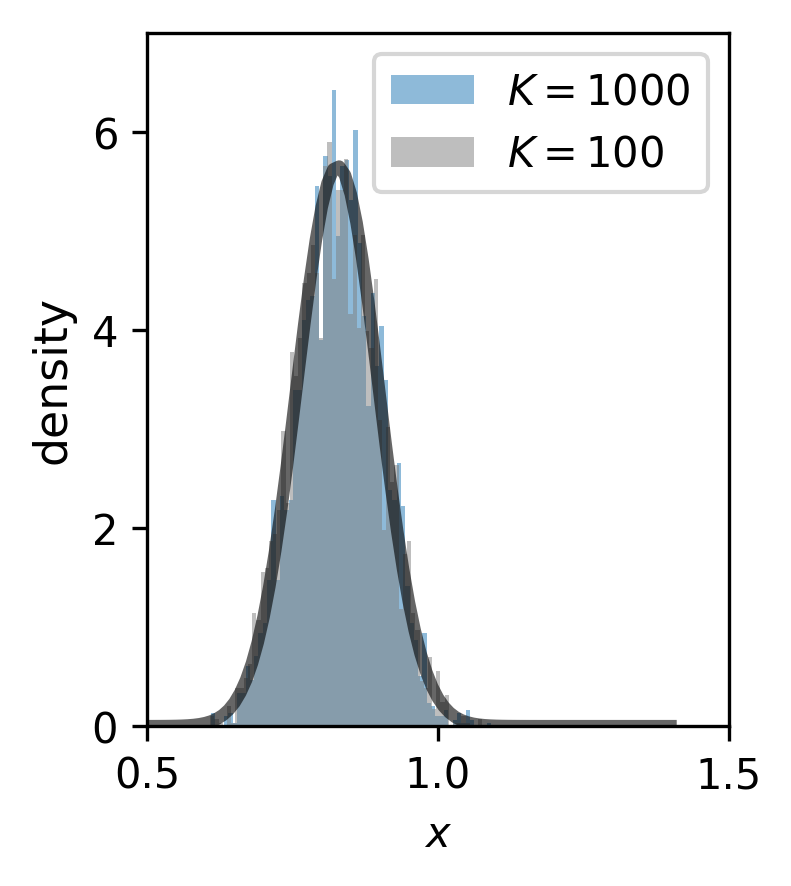}
       \vspace{-5pt}
        \caption{Two-stage MH}
    \end{subfigure}
    \begin{subfigure}[b]{0.22\linewidth}
        \includegraphics[scale=0.51]{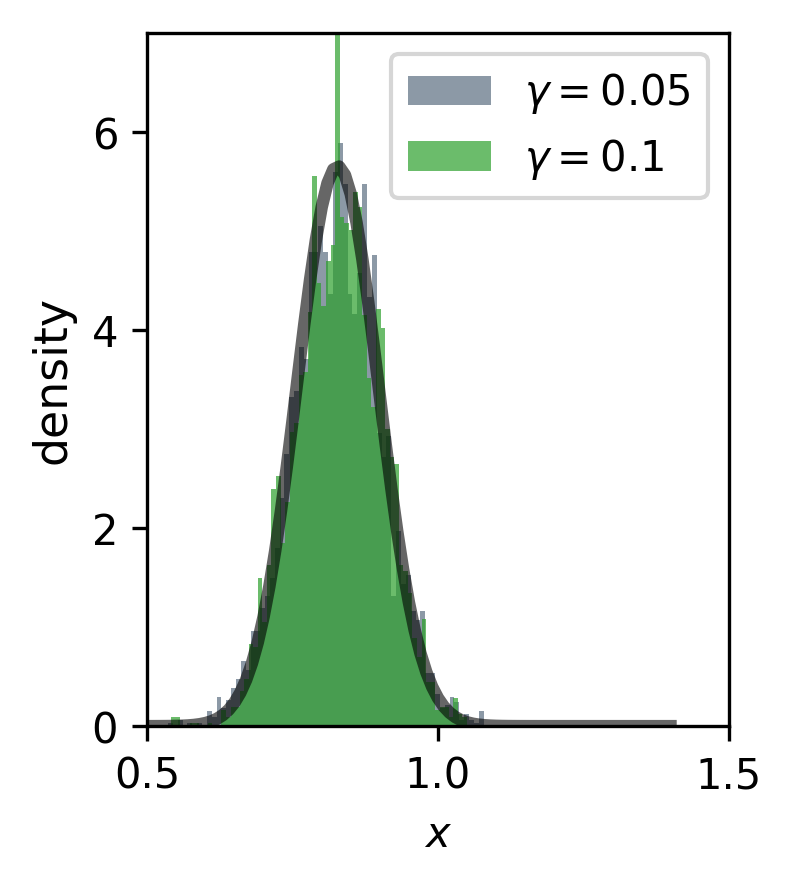}
       \vspace{-5pt}
        \caption{Multi-fidelity MH}
    \end{subfigure}
    \begin{subfigure}[b]{0.43\linewidth}
        \centering
        \includegraphics[scale=0.51]{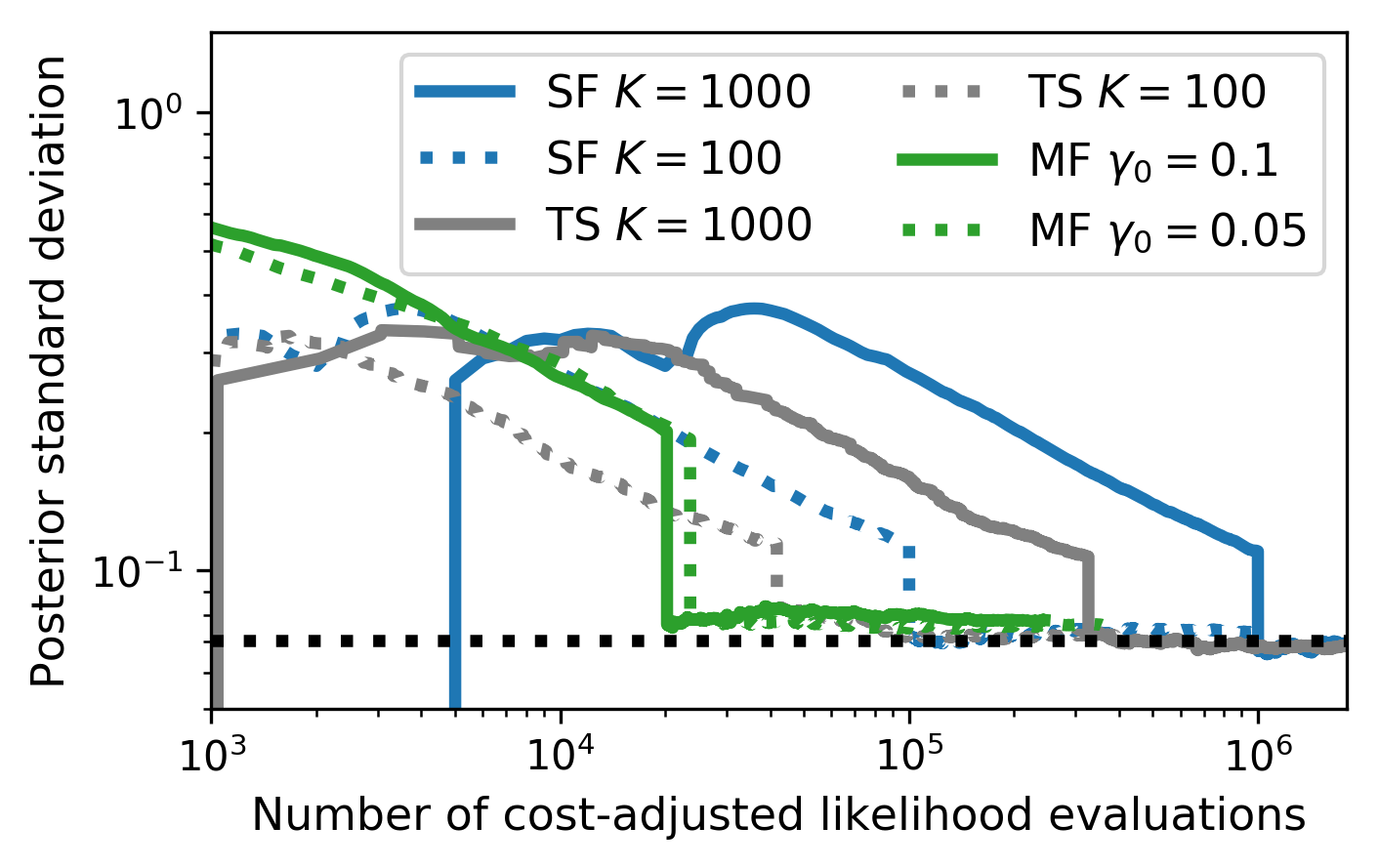}
       \vspace{-5pt}
        \caption{MH sampler estimate vs cost}
    \end{subfigure}
    \begin{subfigure}[b]{0.22\linewidth}
        \centering
        \includegraphics[scale=0.51]{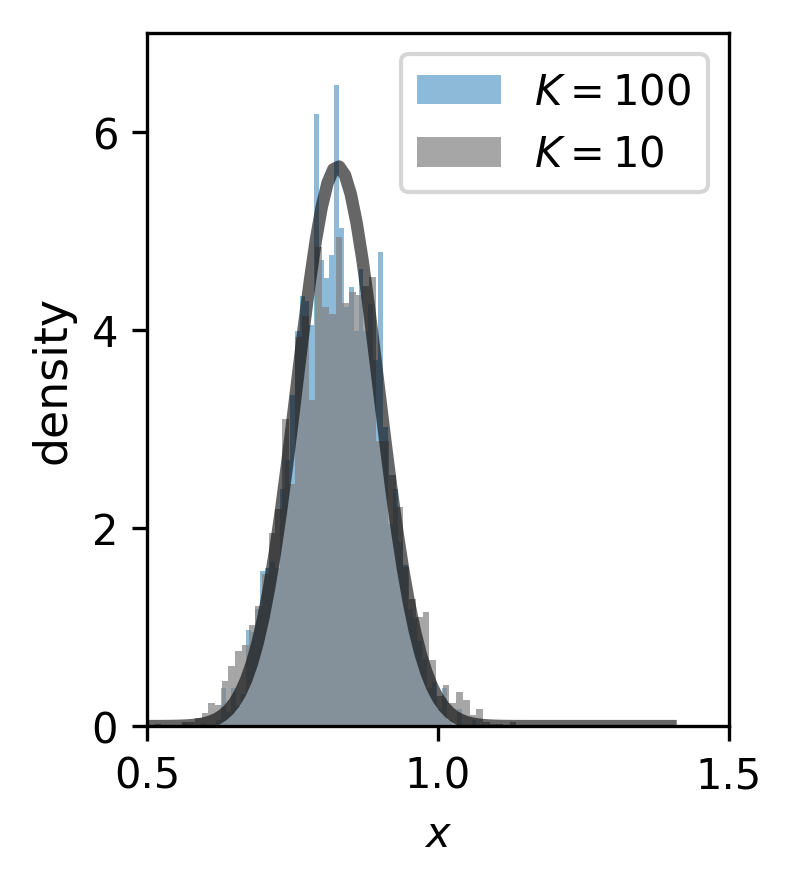}
       \vspace{-5pt}
        \caption{Single-fidelity slice}
    \end{subfigure}
    \begin{subfigure}[b]{0.22\linewidth}
        \includegraphics[scale=0.51]{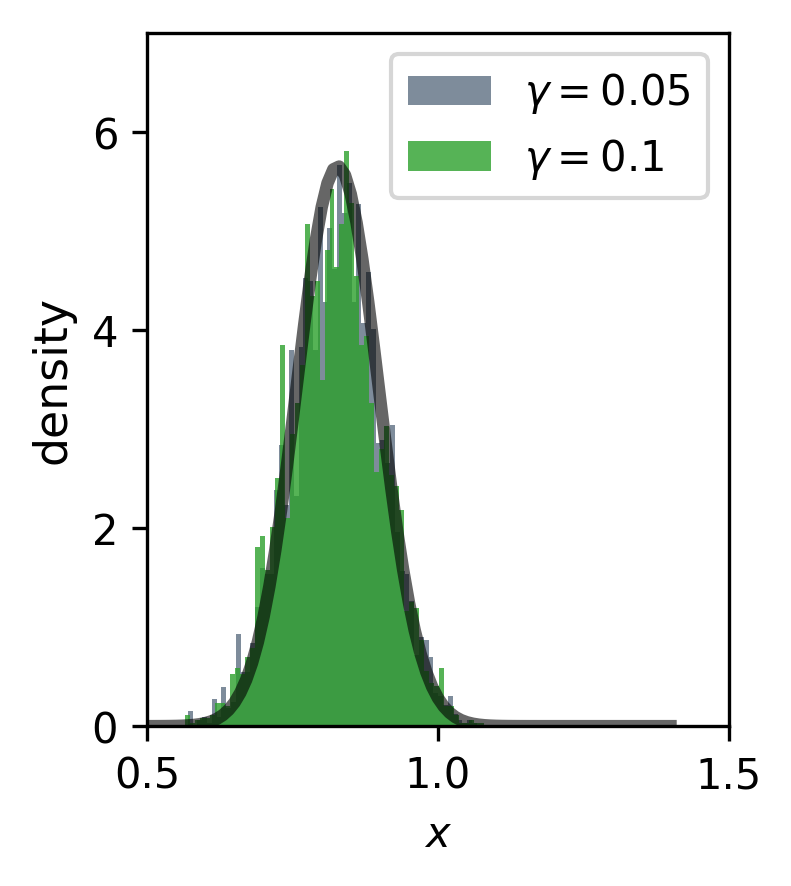}
       \vspace{-5pt}
        \caption{Multi-fidelity slice}
    \end{subfigure}
    \begin{subfigure}[b]{0.43\linewidth}
        \centering
        \includegraphics[scale=0.51]{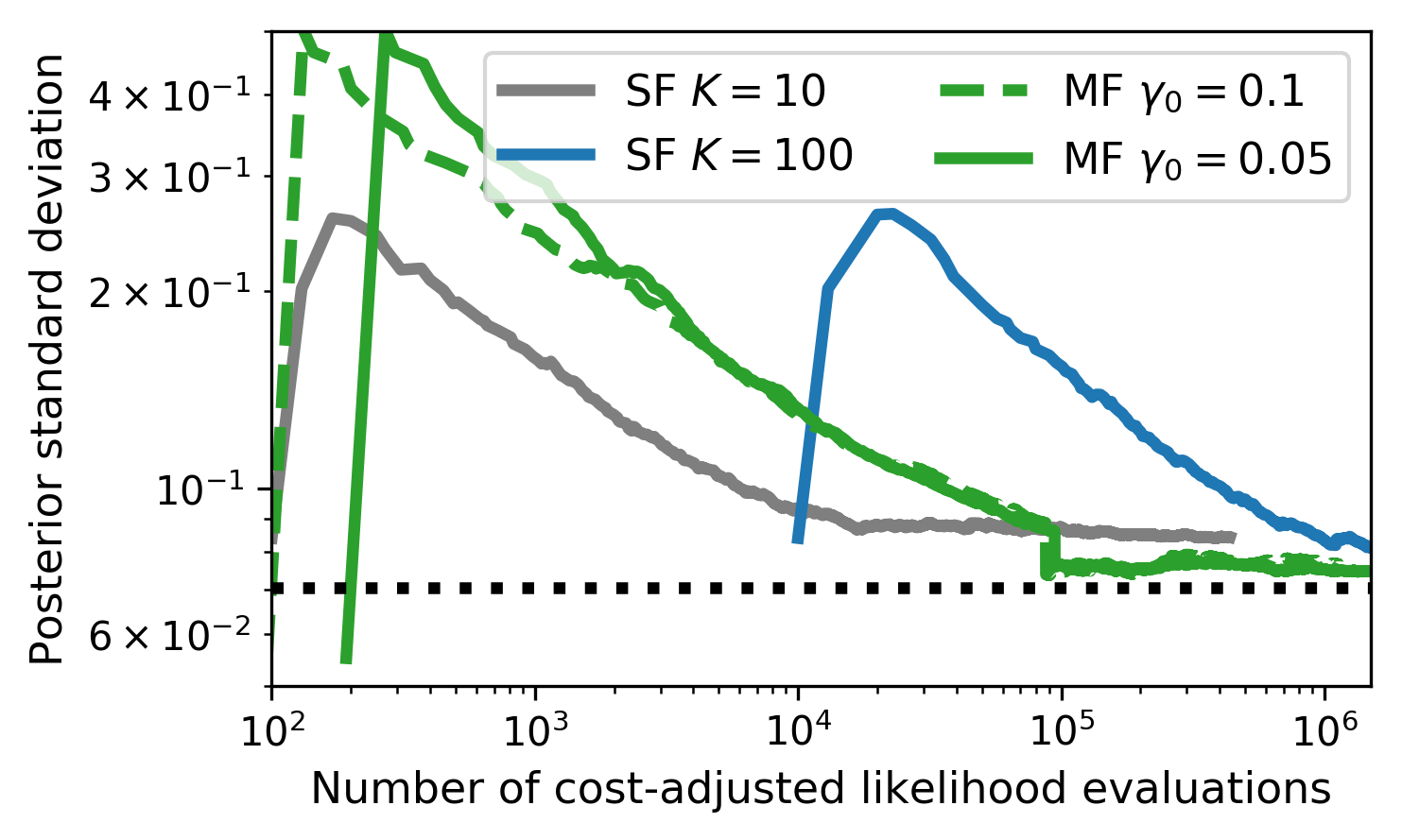}
       \vspace{-5pt}
        \caption{Slice sampler estimate vs cost}
    \end{subfigure}
    \caption{\small%
    Conjugate Gaussian model.
    \textbf{Left:}
    Histograms for M-H (a,b) and slice sampling (d,e).
    \textbf{Right:}
    Comparison of posterior standard deviation estimate vs computation for
    M-H (c) and slice sampling (d) methods.
    Average posterior mean computed over 4 different chains.
     }
    \label{fig:toysequence1}
    \vspace{-2pt}

\end{figure}

\subsection{Toy conjugate Gaussian models}
\label{sec-expt-toy-sequence}

In order to understand the behavior of MF-MCMC on a simple
example of Bayesian inference,
we first examine an example
where the computational cost of evaluating the sequence of
low-fidelity likelihoods does not increase with $k$.
Consider a perfect-fidelity likelihood $L_\infty(\theta) = \mathcal{N}(x; \theta, \sigma_\infty)$
and a low-fidelity likelihood $L_k(\theta) =  \mathcal{N}(x; \theta, \sigma_k)$,
where $\sigma_k^2 \rightarrow \sigma_\infty^2$.
The prior is $\pi_0(\theta) = N(\theta | 0, 1)$, and so a closed-form posterior density can be computed.
Here  we consider the sequence $\sigma_k^2 = 1 + 2/k^2$ and $\sigma_\infty^2 = 1$.
In \Cref{fig:toysequence1} we compare the results of single-fidelity and multi-fidelity M-H and
slice sampling as well as
the two-stage M-H algorithm summarized in \Cref{appendix:twostage}
Here we
consider 2 two-stage M-H with high and low fidelities of of $\{k^{\text{HF}}, k^{\text{LF}}\} = \{1000,10\}$ and $\{k^{\text{HF}}, k^{\text{LF}}\} = \{100,5\}$.
The histograms show the bias of each method after simulating 10,000 samples, and
the solid gray curve denotes the exact posterior density.
We also compute a measure of total cost and a running
average of the estimate of the posterior standard deviation functional,
where the dotted black line denotes the true value.
The number of cost-adjusted likelihoods was computed by upweighting each likelihood
evaluation by the fidelity.
Here the multi-fidelity methods typically converges to
a similar value as the single high-fidelity methods
but in fewer cost-adjusted likelihood evaluations.

\subsection{Log-Gaussian Cox processes}

\begin{figure}[t]
    \centering
    \includegraphics[scale=0.47]{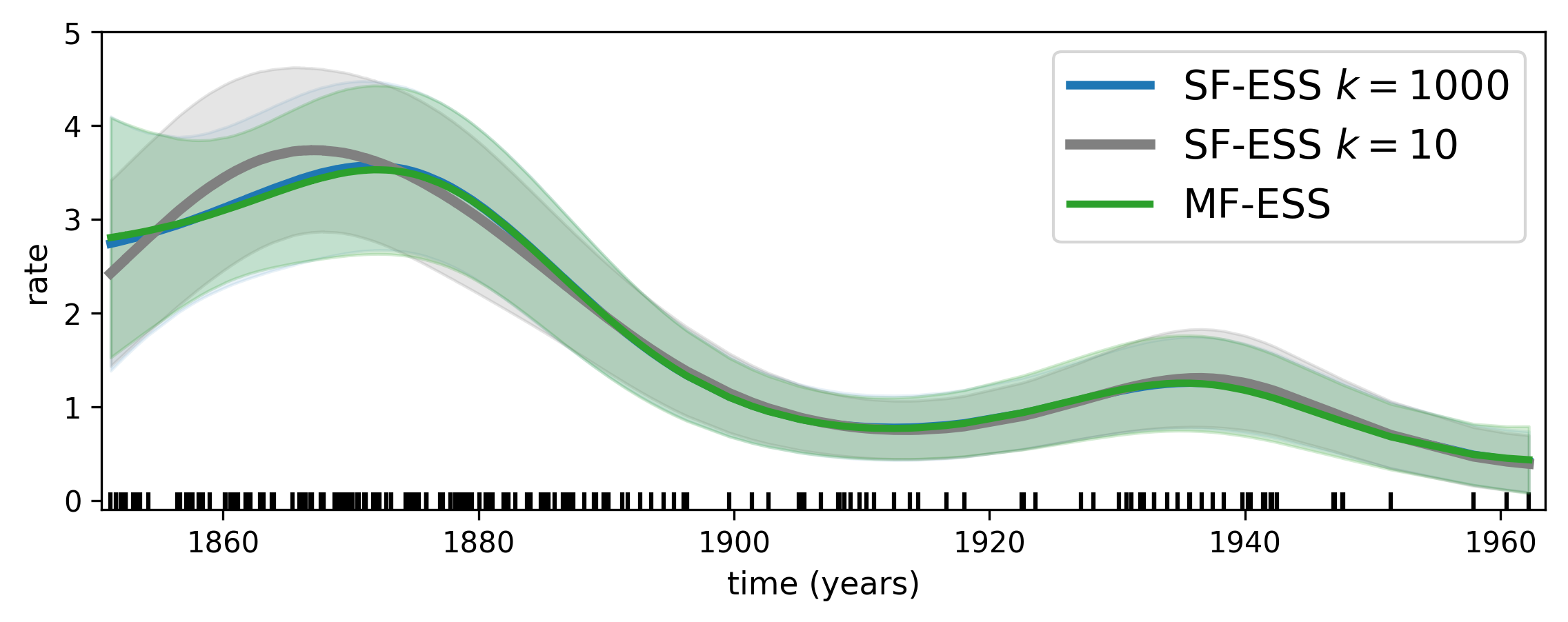}
    \includegraphics[scale=0.64]{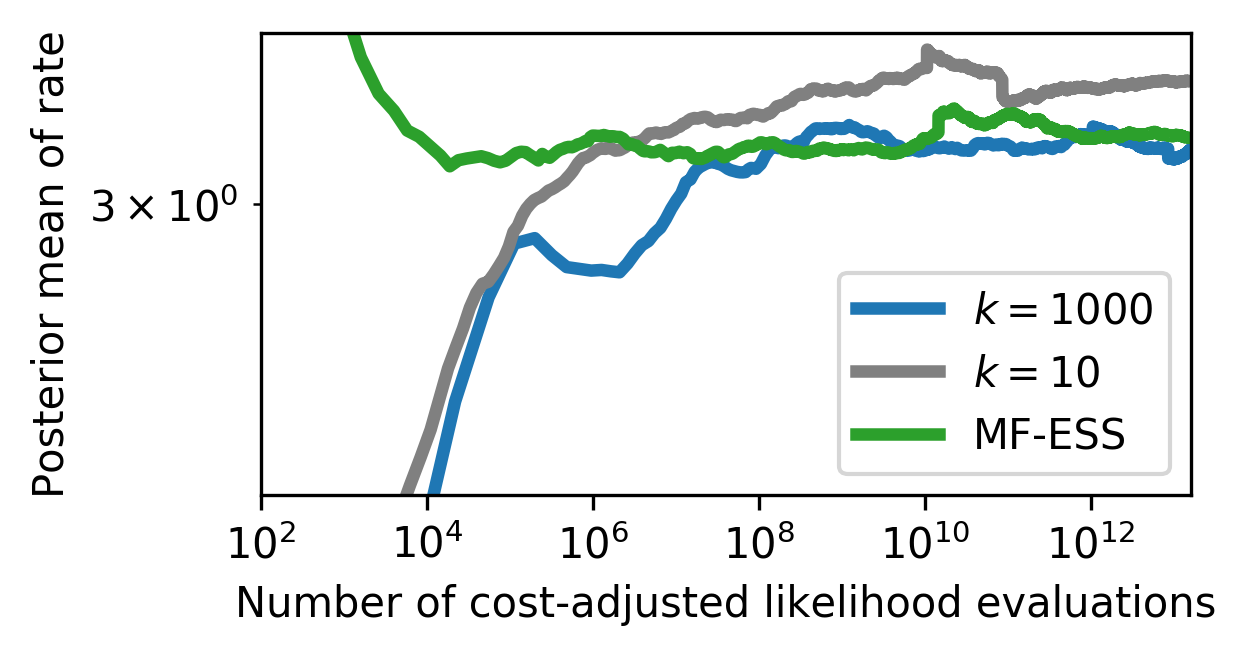}
    \caption{\small Coal mining disasters 1850--1963.
    \textbf{Left:}
    Posterior mean of the rate function at the observed data points.
    \textbf{Right:}
    Posterior mean of the rate function at $T=1862$ vs computation.
    }
    \label{fig:coal}
\end{figure}

We examine an application of MF-MCMC to the log Gaussian Cox process
(LGCP) model \citep{moller1998log}, where the
perfect-fidelity model is a function of an integral and the
lower-fidelity sequence of models arises from $k$-point quadrature estimates.
Let $\log f \sim \GP(0, k_\ell)$, where
$k_\ell(x, x') = \exp\left(-\frac{1}{2\ell^2}\norm{x-x'}_2^2\right)$
and where $\ell$ is a lengthscale hyperparameter.
Consider an inhomogenous Poisson process on $\mathbb{X} \subseteq \reals^D$
with intensity~$\lambda(x) = e^{f(x)}$.
Given a random set of points $\{X_n\}_{n=1}^N$,
the perfect-fidelity likelihood is
\begin{align}
    \label{eq-lgcp}
    L_\infty(f) =
    \exp\left(\int_{\mathbb{X}} (1-e^{f(x)}) dx\right)
    \prod_{n=1}^N e^{f(X_n)}.
\end{align}
Typically, inference in the LGCP uses a grid-based approximation of
\Cref{eq-lgcp}, where the points are binned into counts and modeled with
a Poisson likelihood
\citep{murray2010elliptical,diggle2013spatial,teng2017bayesian},
resulting in in a biased posterior.
Because the likelihood depends on a high-dimensional latent Gaussian vector,
we perform inference for $f$ using the elliptical slice sampling (ESS) algorithm
(see \Cref{appendix-ssec-ess}).
We approximate the integral in \Cref{eq-lgcp} with a trapezoidal quadrature rule
$I_k$, where the number of quadrature points is a linear function of $k$.

We apply multi-fidelity and single-fidelity ESS algorithms to a coal mining disasters
data set (\citet{carlin1992hierarchical}).
The data contain the dates of 191 coal mine explosions that killed ten or
more men in Britain between March 15, 1851 and March 22, 1962.
\Cref{fig:coal} (left) shows the estimated mean intensity and standard deviation on coal mining disasters
data between one run of multi-fidelity ESS and two single-fidelity ESS
runs on a high-fidelity ($k=1000$ quadrature points) and low-fidelity ($k=10$
quadrature points) setting.
In this plot, the high- and multi-fidelity means and standard deviation estimates
match well, and the bias in the low-fidelity estimate is quite noticable.
We also computed the cost-adjusted number of likelihood evaluations performed in each
iteration of MF-ESS and SF-ESS.
\Cref{fig:coal} (right)
shows the average estimated mean intensity at the time step $t=1862$
on the three models against the average cost-adjusted number of
likelihood evaluations per iteration.
Here we observe that the multi-fidelity and high-fidelity estimates are
quite close after many iterations of sampling, but that the multi-fidelity
estimate converges with less computation.

\subsection{Bayesian ODE system identification}
\label{ssec-bayes-ode}

\begin{figure}[t]
    \centering
    \begin{subfigure}[b]{\linewidth}
        \centering
        \includegraphics[scale=0.48]{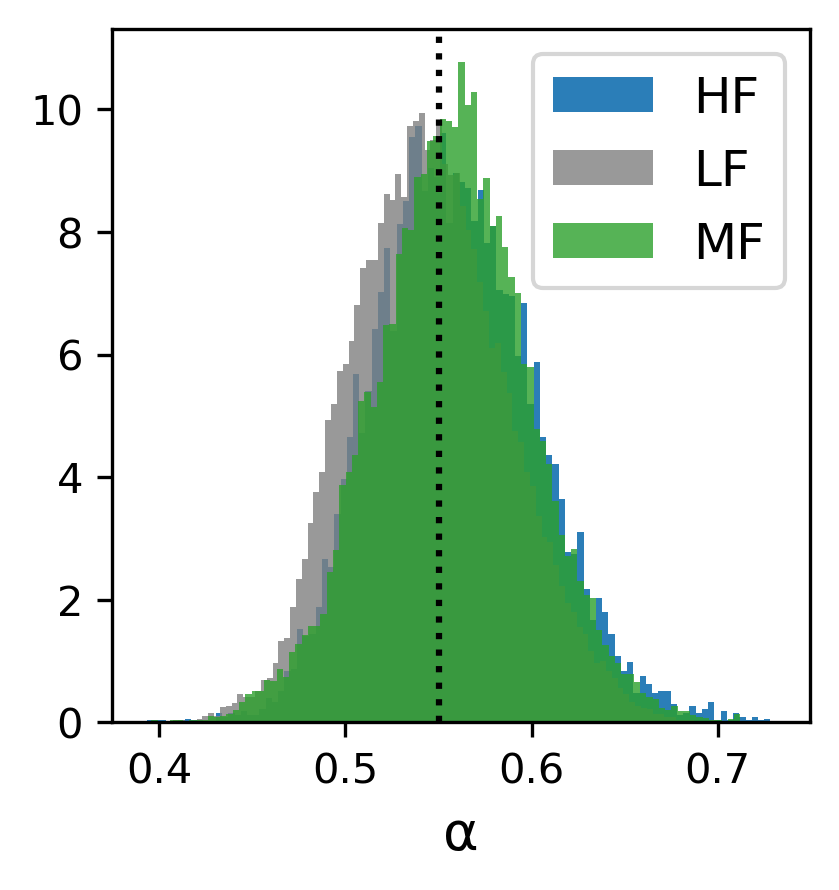}
        \includegraphics[scale=0.48]{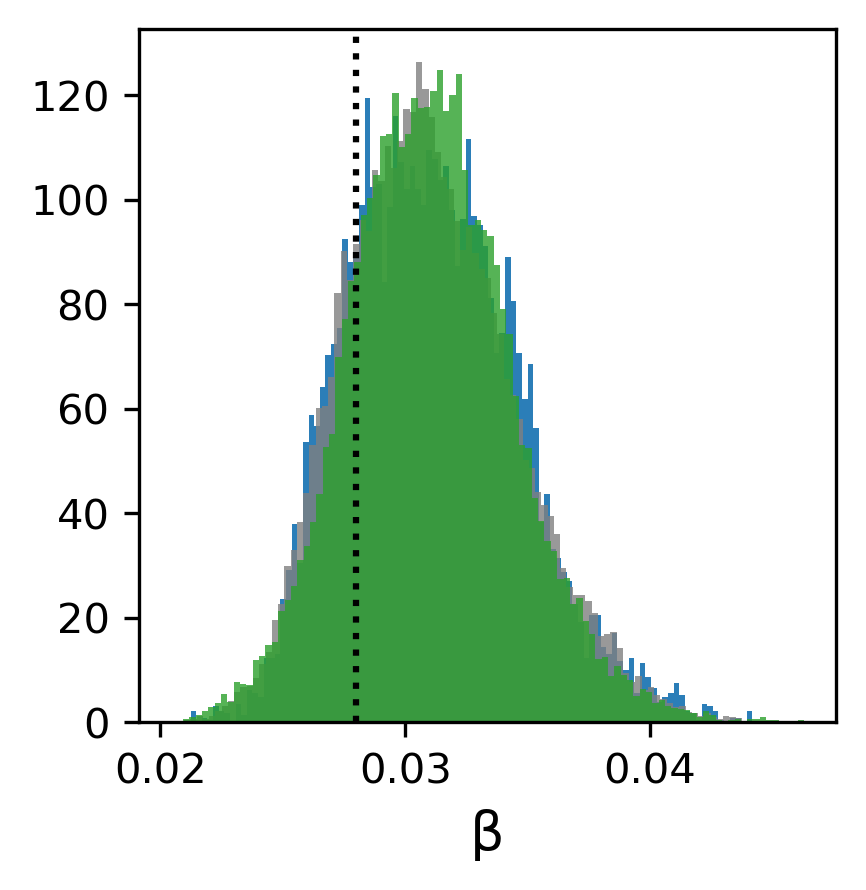}
        \includegraphics[scale=0.48]{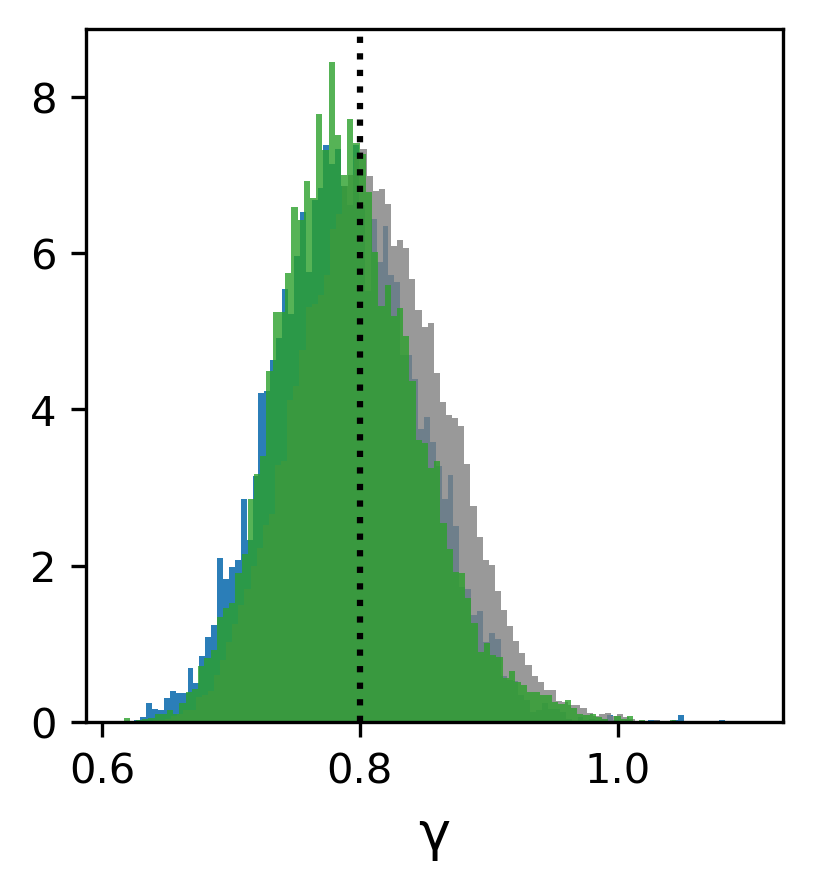}
        \includegraphics[scale=0.48]{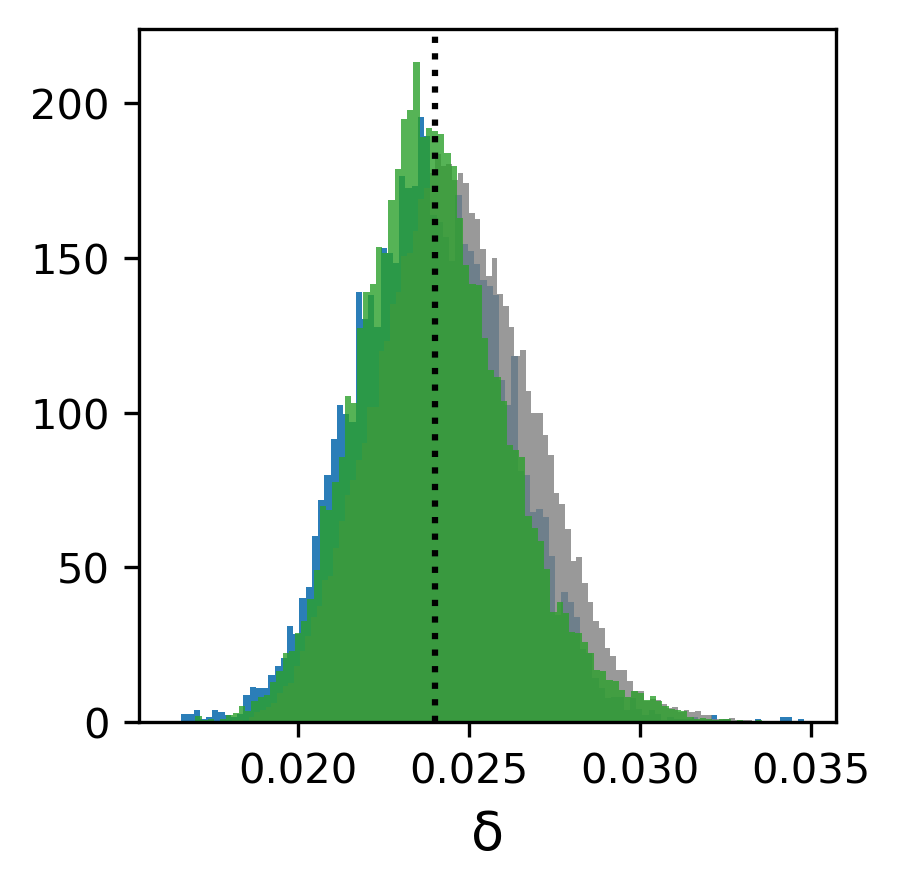}
        \vspace{-5pt}
        \caption{Marginal densities of system parameters}
    \end{subfigure}
    \begin{subfigure}[b]{\linewidth}
       \centering
       \includegraphics[scale=0.66]{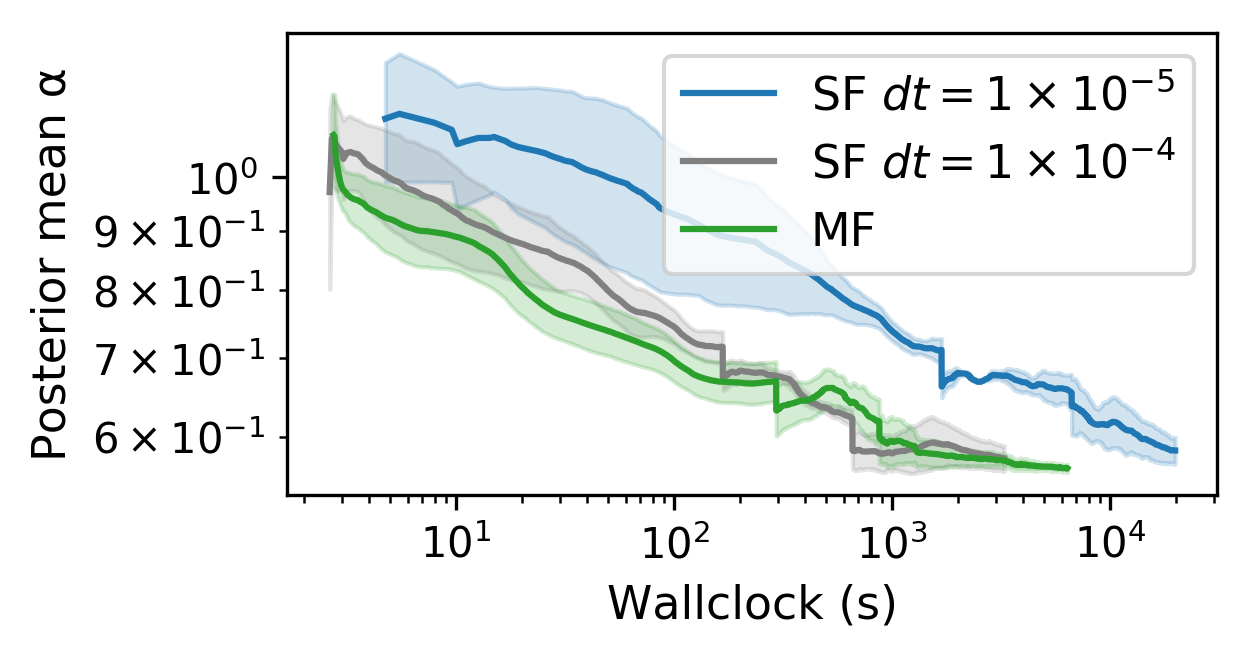}
       \includegraphics[scale=0.66]{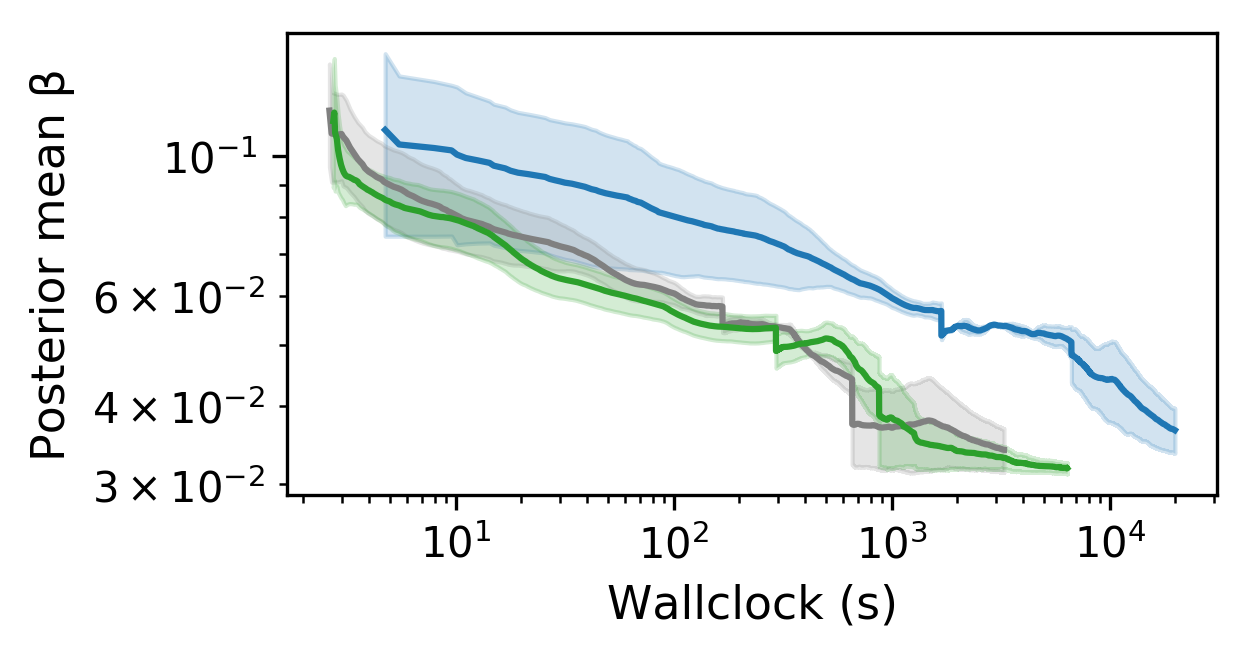}
       \vspace{-5pt}
       \includegraphics[scale=0.66]{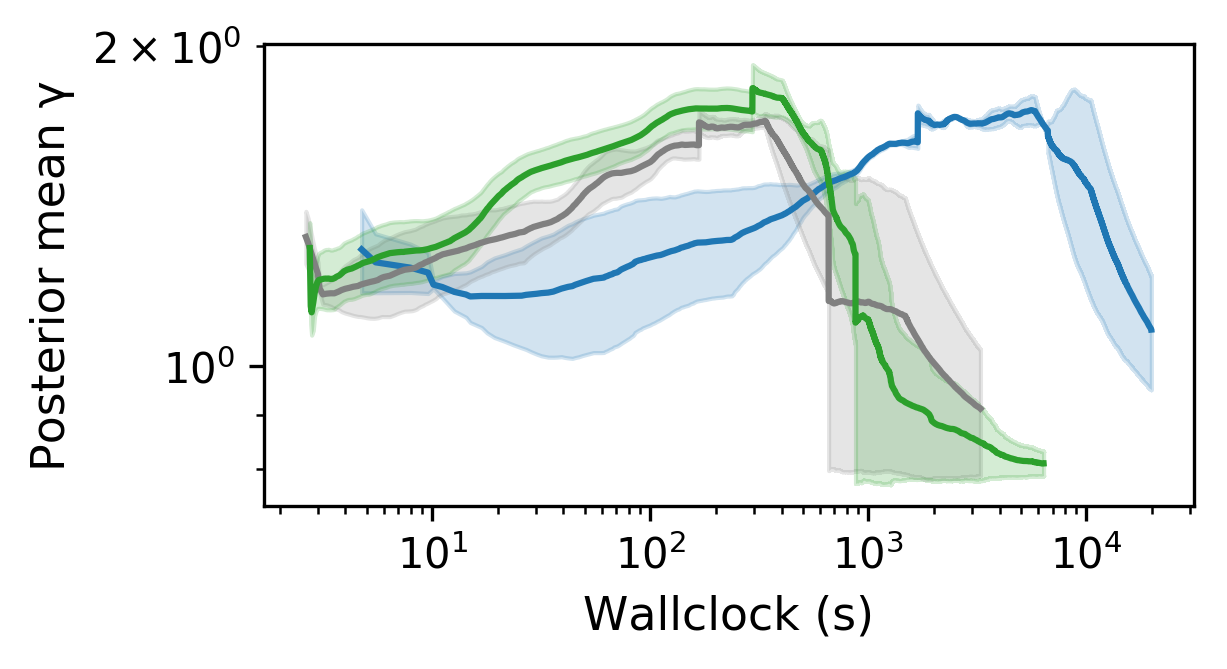}
       \includegraphics[scale=0.66]{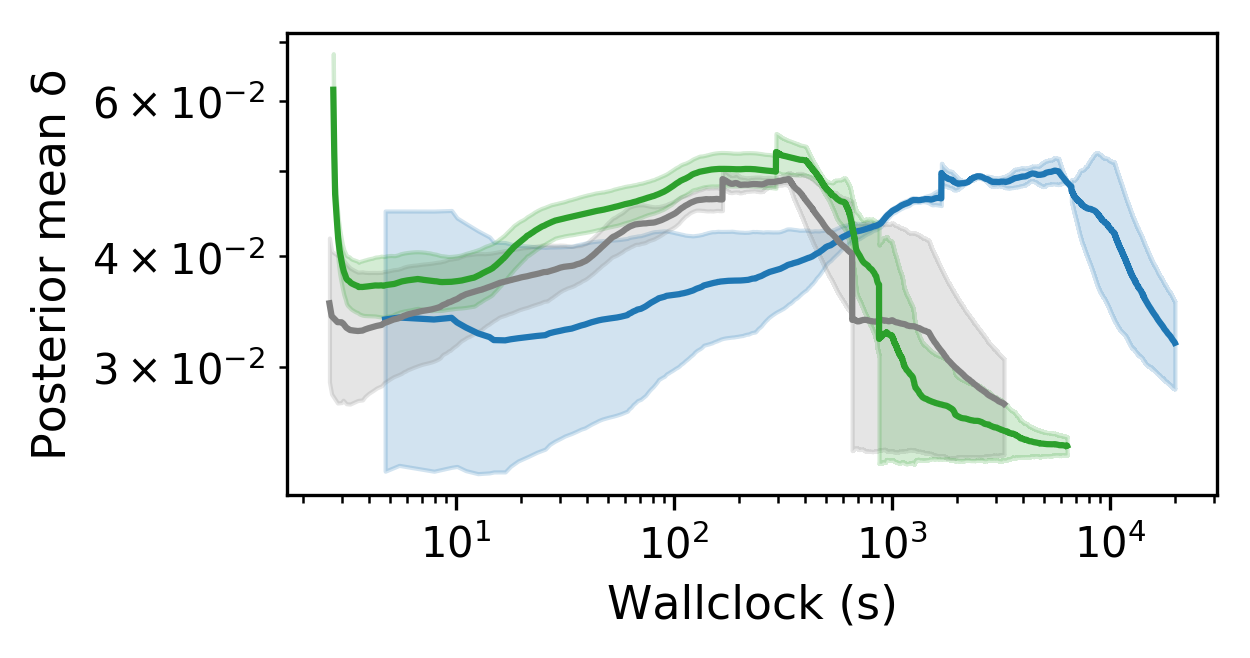}
       \caption{Posterior mean estimate vs computational cost}
    \end{subfigure}
    \caption{\small Lokta-Volterra system parameter identification.
    The fidelity represents (a function of) the step size $dt$ of the ODE solver.
    \textbf{Top:} Marginal distributions of system parameters.
    \textbf{Bottom:}
    Posterior mean estimates of the parameters vs wallclock.
    }
    \label{fig:lvode}
    \vspace{-3pt}
\end{figure}

We now apply the MF-MCMC approach to Bayesian system identification.
Consider the classical Lotka-Volterra ODE problem.
Let
$u(t) \geq 0$ represent the population size of the prey species at time $t$, and
$v(t) \geq 0$ represent the population size of the predator species.
The dynamics of the two species given parameters $\alpha, \beta, \gamma, \delta \geq 0$
are given by a pair of first-order ODEs:
\begin{align}
    \frac{d}{dt} u = (\alpha - \beta v) u = \alpha u - \beta uv, \qquad
    \frac{d}{dt} v = (-\gamma - \delta u) v = -\gamma v - \delta uv.
\end{align}
System identification solves the inverse problem by identifying the parameters of the ODE system
$\theta = (\alpha, \beta, \gamma, \delta)$.
Taking a Bayesian approach, we specify a noise model for the observed data and priors on the
parameters, and we use MCMC to infer a distribution over the solution.
For simplicity, we assume that the initial conditions are known and fix $\sigma=0.25$.

Define $z_n := (u(t_n), v(t_n))$ and let
$z_1(\theta), \ldots, z_N(\theta)$ be the solutions to the Lotka-Volterra differential equations
at times $t_1,\ldots,t_N$ given the initial conditions
and the system parameters~${\theta = (\alpha, \beta, \gamma, \delta)}$.
Suppose we have observations arising from~${\log(y_{n}) = \log(z_n) + \epsilon_n}$, where ${\epsilon \sim N(0, \sigma^2 I)}$.
The low-fidelity likelihood is a function of a numerical solution of the ODE using a time step of size $dt$
(\Cref{eq-ode-lowfi}).
We compared the performance of a multi-fidelity elliptical slice sampler to two
single-fidelity (HF and LF) ellipitical slice samplers with step size $dt = 1\times 10^{-5}, 1 \times 10^{-4}$.
For the ODE solver, we considered both an Euler solver and an 4th-order Runge-Kutta solver.
Results for the Euler solver are in \Cref{fig:lvode}, and results for the Runge-Kutta sovler are in
\Cref{fig:lvode1}.
In the posterior mean estimates of the parameters,
the estimates from the multi-fidelity slice sampler converge in
less time than the single-fidelity samplers.

\subsection{PDE-constrained optimization}

\begin{figure}[t]
    \centering
    \begin{subfigure}[b]{0.24\linewidth}
        \centering
        \includegraphics[scale=0.26]{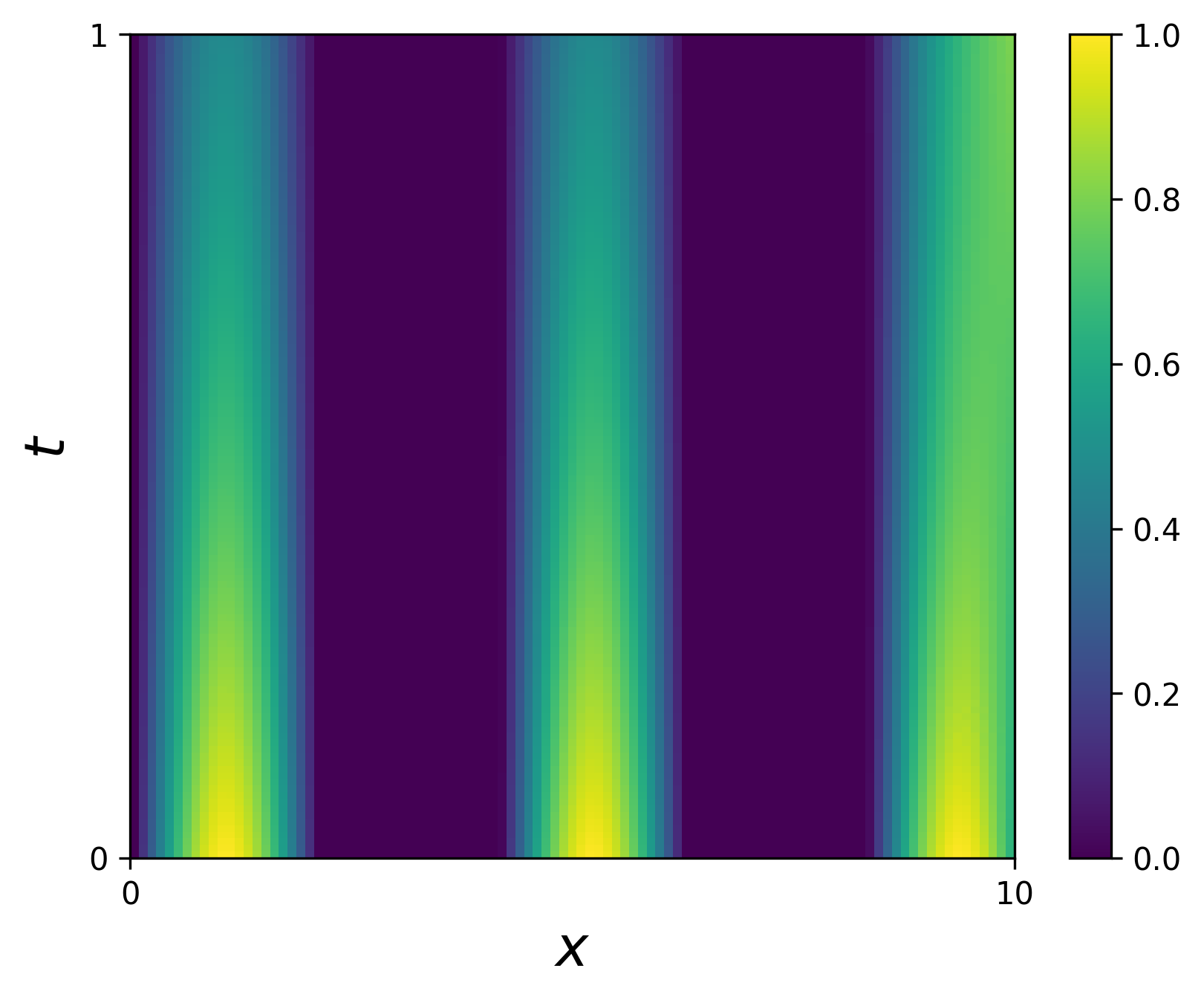}
        \caption{Target temperature $\bar{u}$}
    \end{subfigure}
    \begin{subfigure}[b]{0.24\linewidth}
        \centering
        \includegraphics[scale=0.26]{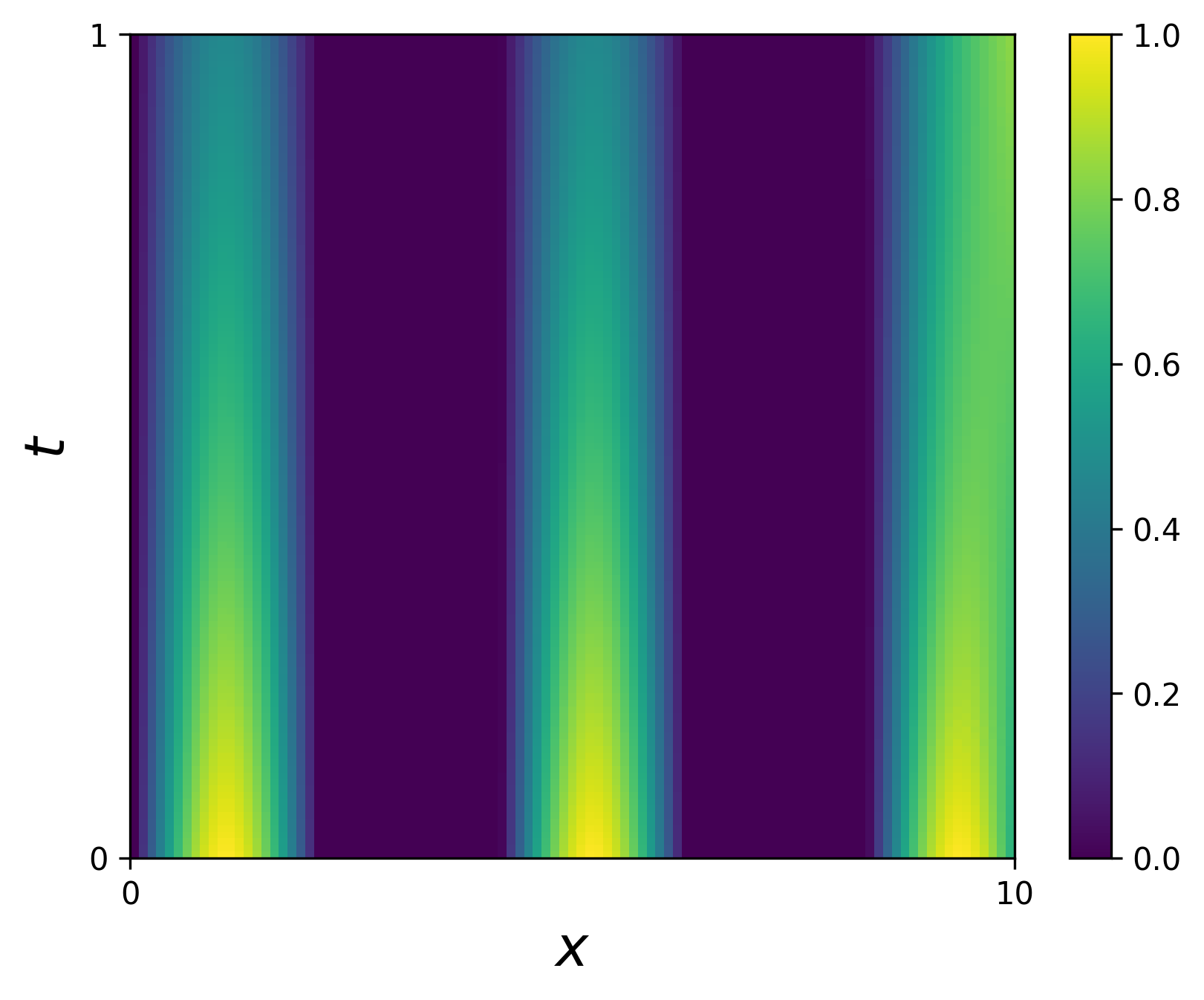}
        \caption{Multi-fidelity solution}
    \end{subfigure}
    \begin{subfigure}[b]{0.24\linewidth}
        \centering
        \includegraphics[scale=0.26]{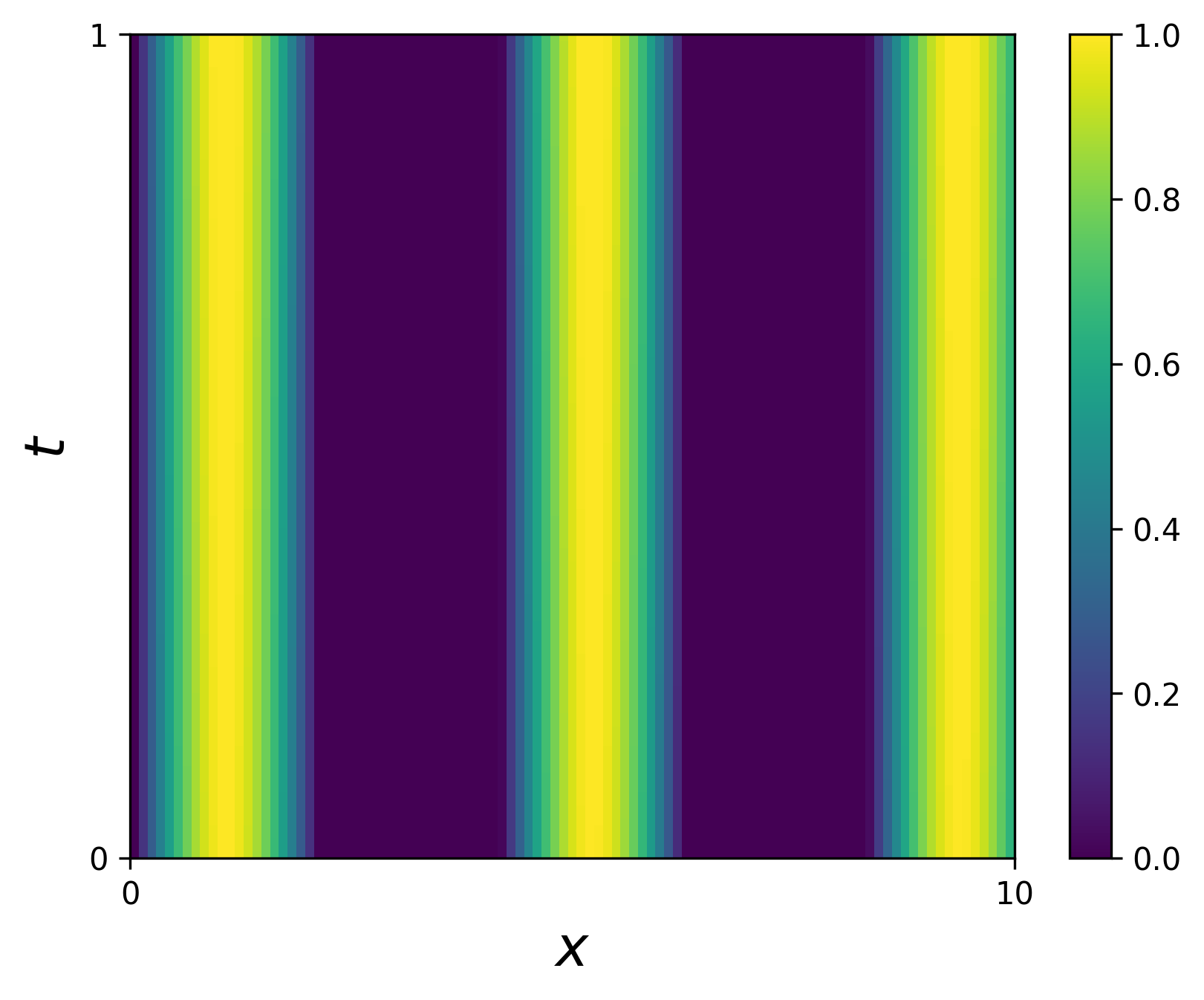}
        \caption{Low-fidelity solution}
    \end{subfigure}
    \begin{subfigure}[b]{0.24\linewidth}
        \centering
        \includegraphics[scale=0.26]{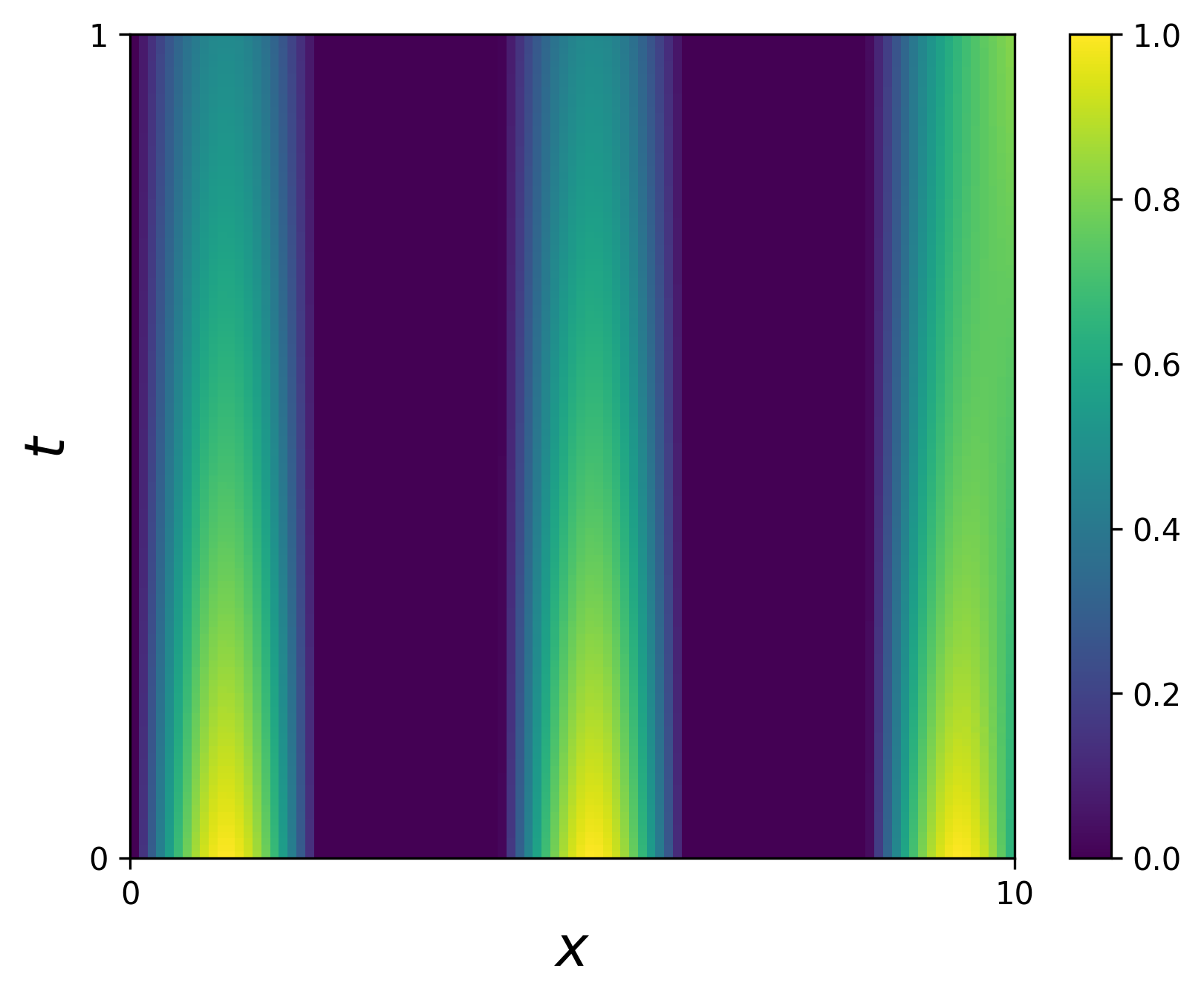}
        \caption{High-fidelity solution}
    \end{subfigure}
    \begin{subfigure}[b]{\linewidth}
       \centering
       \includegraphics[scale=0.73]{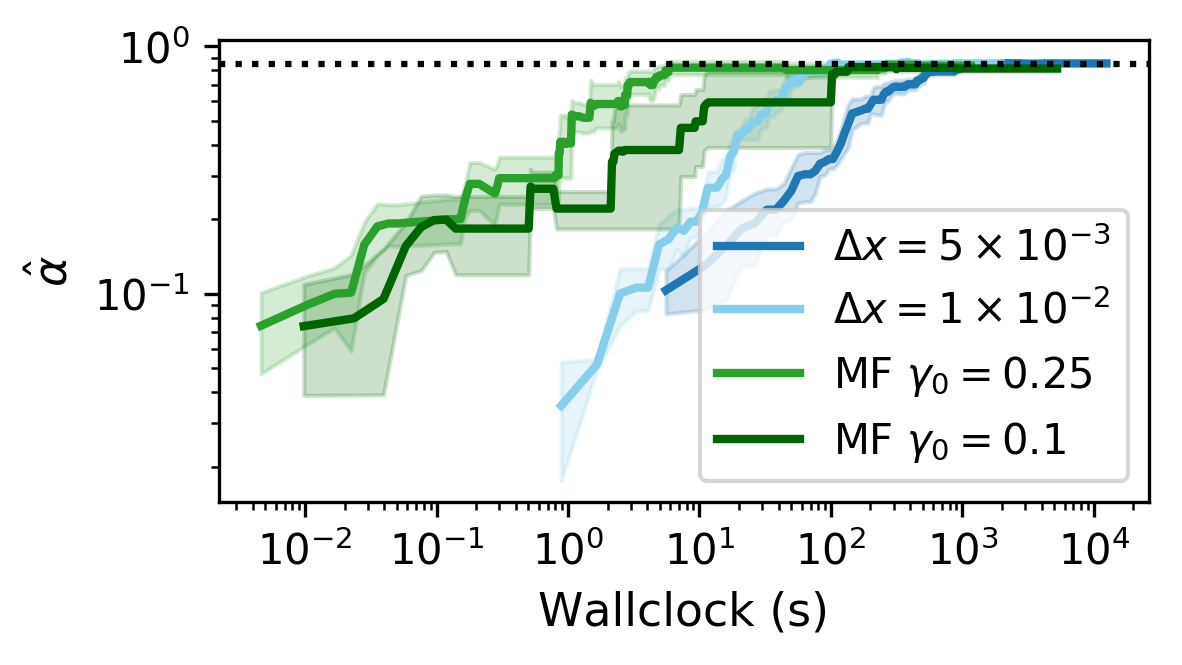}
       \includegraphics[scale=0.73]{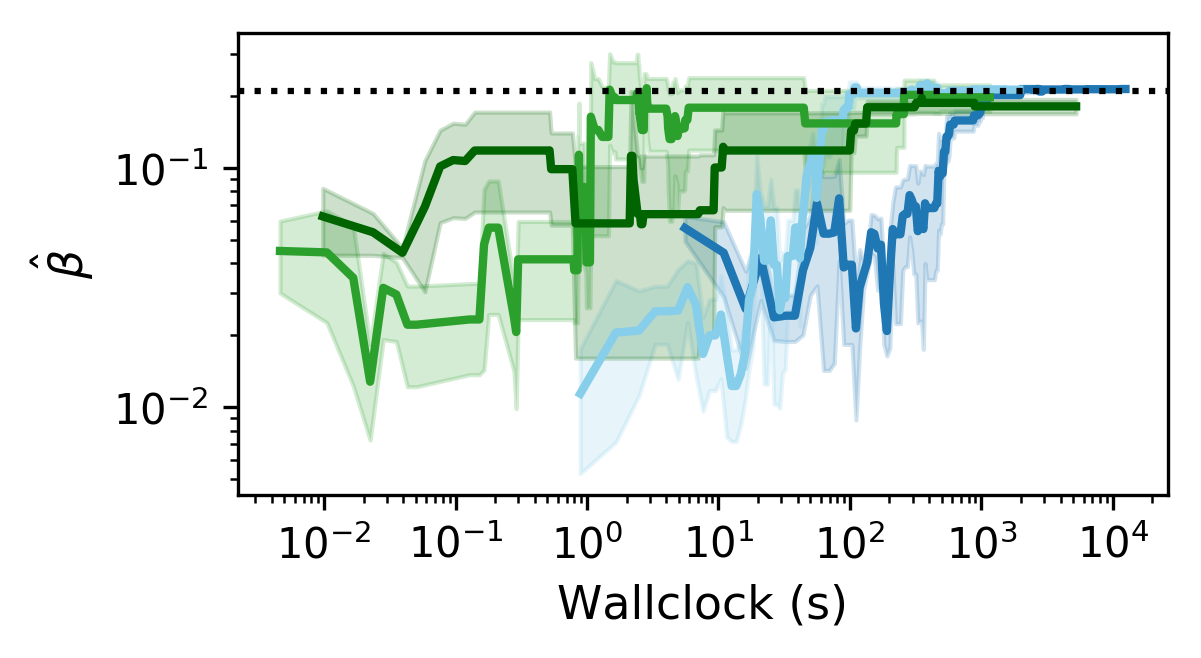}
        \caption{
        Estimate for $\alpha$ (left) and $\beta$ (right) vs computation}
    \end{subfigure}
    \caption{
        \small
        PDE-constrained optimization with a linear heat equation.
        \textbf{Top:} Target and solution temperature functions.
        \textbf{Bottom:} Estimated parameters.
    The black dotted line denotes the true values of $\alpha,\beta$.
    }
    \label{fig:pdeconst}
\end{figure}

We now consider global optimization of a PDE-constrained objective, where an expensive physical simulation
is run repeatedly in an outer loop problem.
A common approach for global optimization is simulated annealing, which has been applied to
constrained global optimization \citep{romeijn1994simulated}.
Consider a model for heat flow in a thin rod of length $L$ with spatial coordinates $x \in [0, L]$.
Let $u(x,t)$ represent the temperature in the rod at position $x$ and time $t$,
and let $\bar{u}$ represent a desired target temperature.
The goal is to minimize a loss function
subject to $u$ satisfying a linear heat equation.
This objective along with an initial condition and homogenous boundary conditions can be
summarized as:
\begin{align}
    \text{minimize}_{u} &\quad \norm{u - \bar{u}}_2^2 \nonumber \\
\text{subject~to~}\qquad
    \frac{\partial u}{\partial t} &= \alpha \cdot \frac{\partial^2 u}{\partial x^2} + 2\beta \cdot u \\
u(x,0) &= \sin\left(\pi x/2\right),
\quad u(0,t) = u(L,t) = 0,
    &\quad x \in [0, L], t \in [0, T]. \nonumber
\end{align}
where $\alpha, \beta > 0$ are the system parameters.
The goal is to find $\theta = (\alpha,\beta)$ that minimizes the objective and satisfies the constraints.
To solve the PDE, we discretize the domain into a grid of size $\Delta x$
and represent the  second derivative using the central difference formula. This
induces a system of ODEs that we solve numerically using  a Tsitouras 5/4 Runge-Kutta method,
setting $\Delta t = 0.4  \Delta x^2$ so as to satisfy a CFL stability condition.
Here the fidelity of the problem is given by the size of the spatial discretization $\Delta x$,
which in turn controls $\Delta t$.
We compared against two single-fidelity discretizations of the
spatial coordinate, where $\Delta x = 5 \times 10^{-3}, 1\times 10^{-2}$.
The results are in \Cref{fig:pdeconst}, where
we plot two of the MF results with $\gamma_0 = 0.1,0.25$.
In these examples, the multi-fidelity estimates converge faster than the
single-fidelity estimates in wallclock time.

\subsection{Gaussian process regression parameter inference}

\begin{figure}[t]
    \centering
    \begin{subfigure}[b]{0.53\linewidth}
        \centering
        \includegraphics[scale=0.58]{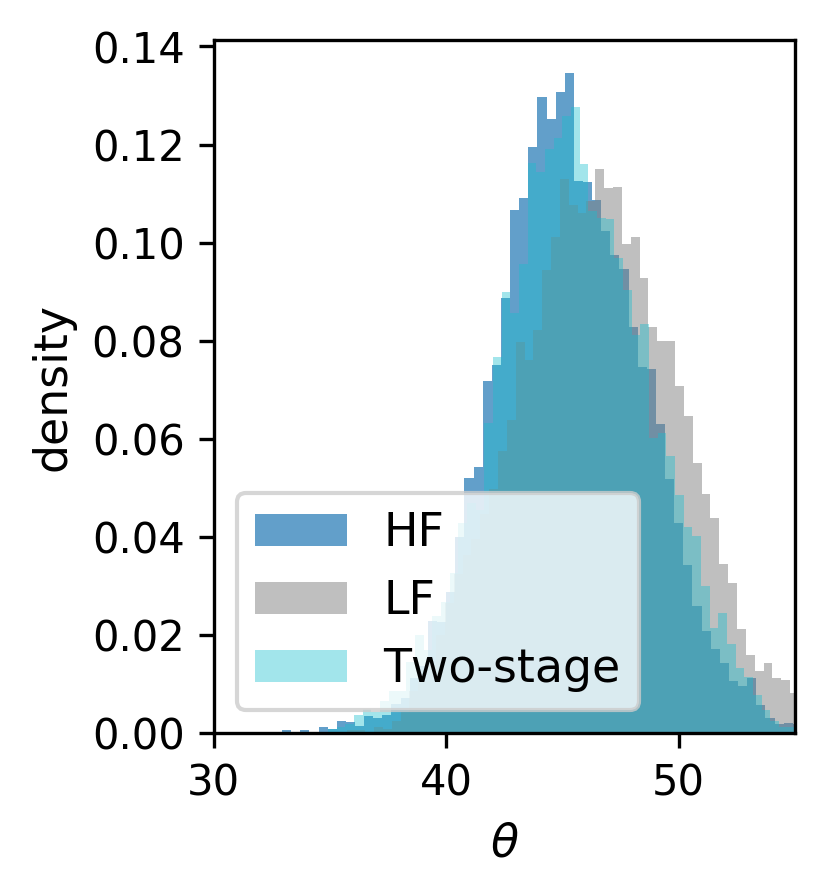}
        \includegraphics[scale=0.58]{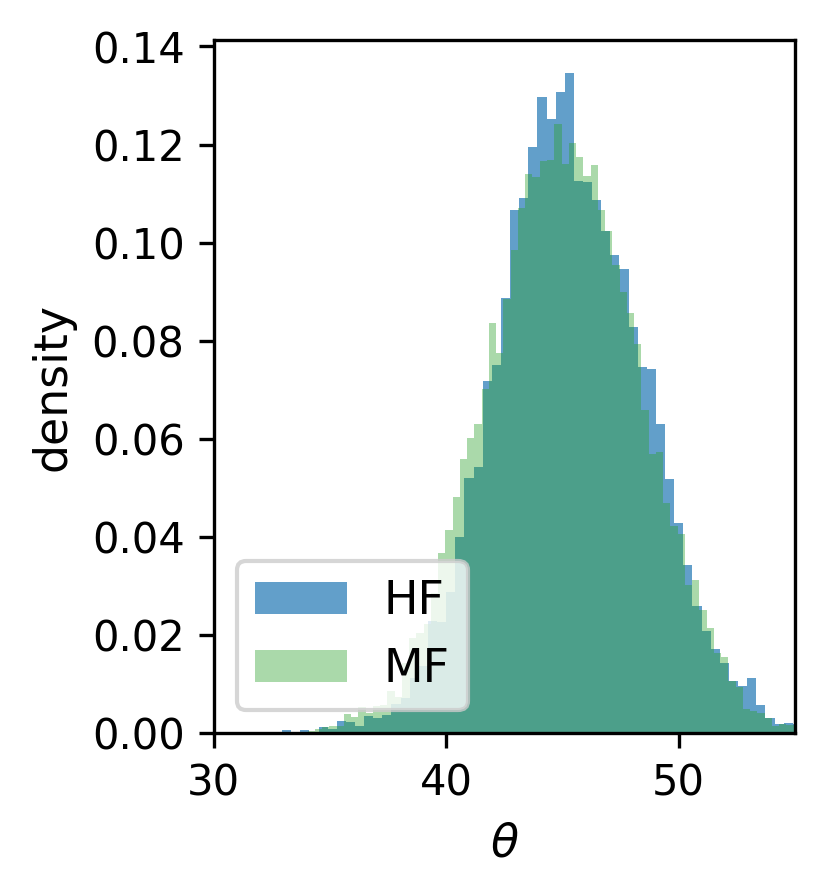}
        \caption{Histograms of high, low, two-stage, and multi-fidelity}
    \end{subfigure}
    \begin{subfigure}[b]{0.45\linewidth}
        \centering
        \includegraphics[scale=0.58]{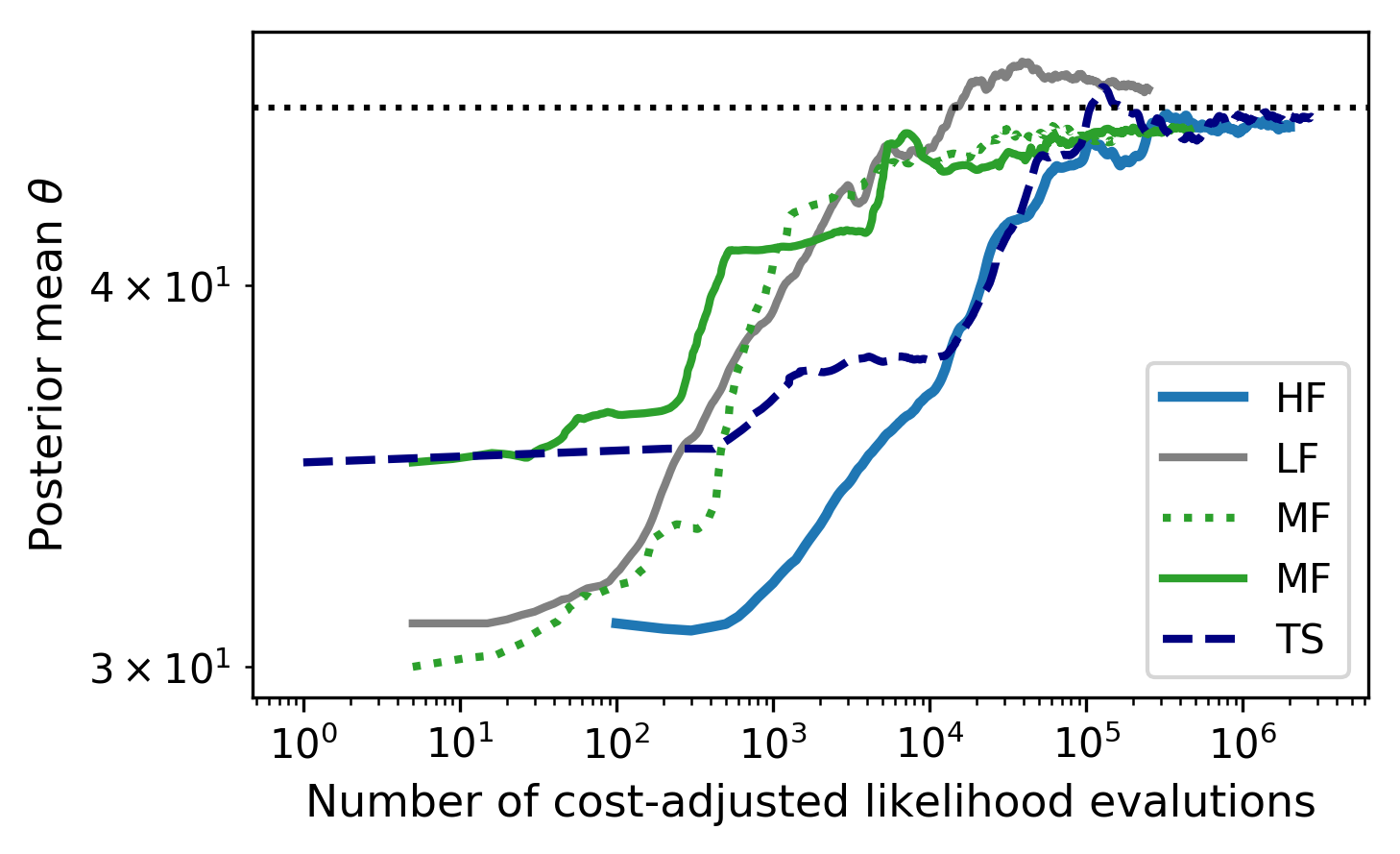}
        \caption{Posterior mean estimate vs computation}
    \end{subfigure}
      \caption{
          \small
        Parameter inference in a Gaussian process regression model.
        \textbf{Left:} The posterior distribution of the parameter $\theta$.
        \textbf{Right:} The posterior mean estimate vs computional cost.
    }
    \vspace{-5pt}
    \label{fig:GPreg}
\end{figure}

Let $X \in \reals^{N \times D}$ and
consider a Gaussian process regression model with a squared exponential kernel:
\begin{align}
    \label{model-GP}
    f \sim \GP(0, k_{\theta}),
    \quad
    y = f(X) + \epsilon,
    \quad \epsilon \sim \mathcal{N}(0, \sigma_0^2),
    \quad
    k_\theta(x, x') = \exp\left(-\frac{1}{2\theta^2}\norm{x-x'}_2^2\right).
\end{align}
For simplicity, we assume $\sigma_0^2$ is known.
Let $\Sigma_{\theta} := [k_\theta(x_i, x_j)]_{i \in [N],j \in [N]}$.
In many applications of GPs, e.g., computing the predictive distribution,
 one is interested in integrating out the parameters $\theta$ using MCMC.
Computing the posterior $\pi(\theta | X, y)$ is expensive because each evaluation
of the likelihood $p(y | X, \theta) = \N(y \given 0, \Sigma_{\theta} + \sigma_0^2 I)$ involves computing a determinant
and solving a linear system with respect to the matrix $\Sigma_\theta + \sigma_0^2 I$,
which has an $O(N^3)$ computational cost associated with standard methods (e.g., Cholesky decomposition).
For simplicity, we will only consider the fidelity of solution to the linear
    system, but we note that the determinant can considered using the approach
    described in \citet{potapczynski2021bias}.
Additional derivations and details are in  \Cref{ssec-appendix-gp}.
We generate synthetic data from the GP model with $N=100$,
$\sigma_0^2 = 1$, and lengthscale $\theta_0 = 45$.
For the GP model, we use a log Normal prior on $\theta$ given above in
\Cref{eq:posterior} with parameters $\nu_0 = 3.8, \nu_1 = 0.03$.
We compare several likelihoods:
a high-fidelity likelihood ($K=N$),
low-fidelity likelihood ($K = K_N \ll N$),
and
the multi-fidelity approach we describe with $\gamma_0 = 0.1$.
In \Cref{fig:GPreg}, we compare these approaches.
The low-fidelity likelihood sequence was constructed by computing the solution to the linear system
using a preconditioned conjugate gradient solver with $k$ steps.
Finally, we also compare to a two-stage M-H approach with $k \in \{100,5\}$.
For all methods, we use a M-H sampler with $T=50000$ iterations.
The results are in \Cref{fig:GPreg}.
In the histograms, we observe that the high-fidelity, multi-fidelity
and two-stage approaches tend to lead to similar posteriors,
while the low-fidelity sampler has more noticable bias.
The estimate produced by the multi-fidelity samplers converged
in fewer cost-adjusted likelihood evaluations than the high-fidelity and
two-stage approaches.

\section{Discussion and future work}

In this work, we introduced a class of multi-fidelity MCMC that uses a low-fidelity unbiased
estimator to reduce the computational cost of sampling while still maintaining the
desired limiting target distribution of the Markov chain.
In particular, we have demonstrated the use of our framework on more advanced MCMC algorithms beyond M-H,
such as slice sampling, and to additional settings such as optimization.
Our results  show a reduction in computation while producing accurate solutions in comparison with
high-fidelity models when it is possible to construct a target estimator that is not too noisy.
Many future directions remain.
First, applying MF-MCMC to large-scale expensive applications has many
computational challenges.
Making the method more robust to specialized problems is important, especially if
the estimator is heavy-tailed.
Thus, constructing good proposal distributions matching properties of the low-fidelity target
sequence is crucial, especially for application to high-dimensional problems.
In addition, we have thus far focused on target densities where there is a single computation
whose fidelity is varied.
However, in many settings, there may be
target densities with multiple computations that converge at different rates,
for example, if the target density includes both an intractable integral and a solution of a linear system.
Our framework can be extended to that setting by adjusting the proposal distribution,
and it is useful to understand how these rates impact the convergence properties of the sampler.

\subsection*{Acknowledgments}
This work was partially supported by NSF grants IIS-2007278 and OAC-2118201.
D.\ Cai was supported in part by a Google Ph.D.\ Fellowship in Machine Learning.

\appendix

\counterwithin{figure}{section}
\counterwithin{algorithm}{section}
\counterwithin{equation}{section}

\section{Additional related work}
\label{sec:additionalwork}

Approximate Bayesian computation (ABC) is a related class of methods to pseudo-marginal MCMC for
implicit likelihoods.
We note that the asymptotic target for ABC is an approximation to the target density of interest.
\citet{prescott2020multifidelity} propose a multi-fidelity approach to ABC.
A number of multilevel ABC approaches
\citep{guha2017multilevel,lester2018multi,warne2021multifidelity} have also been proposed in recent
work.

In our work, the Russian roulette estimator is used to construct an unbiased, low-fidelity
likelihood.
The Russian roulette estimator has been used recently in a number of applications for
optimization and inference \citep{beatson2019efficient,luo2020sumo,potapczynski2021bias}.
In particular, \citet{potapczynski2021bias} apply similar techniques to get an unbiased estimate
of the gradient of the marginal likelihood for Gaussian process regression; this can viewed as an
optimization analog to our approach.

\section{Review of MCMC algorithms}

In this section, we review the MCMC algorithms used in the main paper.

\subsection{Random-walk Metropolis-Hastings}

Draw proposal $\theta' \sim q(\cdot \given \theta)$.
Accept or reject the proposed value according to:
\begin{align*}
    R = \min\left(1,
    \frac{\pi(\theta', \mathcal{D}) q(\theta | \theta')}
    {\pi(\theta, \mathcal{D}) q(\theta' | \theta)}
    \right)
\end{align*}

In our experiments, we use a proposal distribution of the form $q(\theta' | \theta) = \N(\theta' | \theta, \tau)$, where the parameter $\tau$
needs to be tuned.

\subsection{Slice sampling}

Slice sampling \citep{neal2003slice} is auxillary-variable algorithm that automatically
generates proposals without the need for an explicit accept/reject step.


Let $P(\theta)$ denote the joint distribution of interest.
Given current state $\theta$, sample a uniform random variable
$u \sim \text{Unif}(0, p(\theta))$.
This random variable induces a height at the current state given by $u' = u P(\theta)$.
A  horizontal bracket $(\theta_l, \theta_r)$ is defined around $\theta$
and a proposal $\theta'$ is generated.
If $P(\theta') > u'$, the proposal is accepted; otherwise the bracket is decreased.
We use the ``stepping out'' and shrinking procedures for generating and shrinking the proposal
bracket, as defined in \citet[Chapter~29.7]{mackay2003information}.


\subsection{Elliptical slice sampling}
\label{appendix-ssec-ess}

Elliptical slice sampling \citep{murray2010elliptical} is used for inference in latent Gaussian models.
Let $f \sim N(0, \Sigma)$ denote the latent $D$-dimesional Gaussian variable of interest,
and consider a likelihood $L(f) = p(\mathcal{D} \given f)$.
The target distribution of interest is the joint distribution
\begin{align}
    \pi^*(f) = \frac{1}{Z} \N(f; 0, \Sigma) L(f),
\end{align}
where $Z$ is the marginal likelihood of the model.
The algorithm is summarized in \Cref{alg:ess}.

\begin{algorithm}[t]
\caption{Summary of the elipitical slice sampling iteration (from \citet[Figure~2]{murray2010elliptical}).}
\label{alg:ess}
\begin{algorithmic}[1]
\State \textbf{Input:} Current state $f$, log-likelihood $L$
\State Choose ellipse $\nu \sim \N(0, \Sigma)$
\State Construct log-likelihood threshold:
    \begin{align*}
        u     & \sim \text{Unif}[0,1] \\
        \log y & = \log L(f) + \log u
    \end{align*}
\State Draw initial proposal, define a bracket
    \begin{align*}
        \theta &\sim \text{Unif}[0, 2\pi] \\
        [\theta_{\text{min}}, \theta_{\text{max}}] &\sim [\theta-2\pi, \theta]
    \end{align*}
\State Define proposal $f' = f\cos\theta + \nu \sin\theta$
    \If{$\log L(f') > \log y$}
    \State Accept proposal $f'$
    \Else
    \State Resize bracket and generate new proposal:
        \If{$\theta < 0$}:
        \State $\theta_{\text{min}} = \theta$
        \Else:
        \State $\theta_{\text{max}} = \theta$
        \EndIf
        \State $\theta \sim \text{Unif}[\theta_{\text{min}}, \theta_{\text{max}}]$
        \State Goto Step 5
    \EndIf
\State \textbf{Output:} New state $f'$
\end{algorithmic}
\end{algorithm}

\subsection{Two-stage Metropolis-Hastings}
\label{appendix:twostage}

The two-stage MH algorithm assumes a single high fidelity likelihood $L^{\text{HF}}$
and low fidelity likelihood $L^{\text{LF}}$.
In each iteration $t$,
a proposal $\theta'$ is generated from the proposal distribution $q(\cdot | \theta^{(t-1)})$.

\textbf{Stage 1:}
The proposal is accepted for the second stage or rejected according to the acceptance probability
\begin{align}
    R^{\text{LF}}(\theta; \theta') = \min\left(1,
    \frac{\pi(\theta') L^{\text{LF}}(\theta') q(\theta | \theta')}
    {\pi(\theta) L^{\text{LF}}(\theta) q(\theta' | \theta)}
    \right),
\end{align}
where $\theta = \theta^{(t-1)}$.
If the proposal is rejected, then the value $\theta^{(t)} = \theta^{(t-1)}$.

\textbf{Stage 2:}
In the second stage, the proposal $\theta'$ is accepted with probability
\begin{align}
    R^{\text{HF}}(\theta; \theta') =
    \min\left(1,
    \frac{\pi(\theta') L^{\text{HF}}(\theta') Q(\theta | \theta')}
    {\pi(\theta) L^{\text{HF}}(\theta) Q(\theta' | \theta)}
    \right),
\end{align}
where the proposal distribution $Q$ satisfies
\begin{align}
    Q(\theta' | \theta) = R(\theta',\theta) q(\theta'|\theta) +
    \left(1 - \int R(\theta,\theta')  q(\theta' | \theta) d\theta'\right) \delta_{\theta}(\theta').
\end{align}
Note that in the algorithm, the integral does not need to be explicitly computed,
since if $\theta = \theta'$, the chain remains at the same value, and if
$\theta \neq \theta'$, then
$Q(\theta'|\theta) = R^{\text{LF}}(\theta; \theta') q(\theta' | \theta)$.
Note that the high-fidelity acceptance probability can be easily computed as
\begin{align}
    R^{\text{HF}}(\theta; \theta') = \min\left(1,
    \frac{L^{\text{HF}}(\theta') L^{\text{LF}}(\theta)}
    {L^{\text{HF}}(\theta) L^{\text{LF}}(\theta')}
    \right).
\end{align}
If the proposal is accepted,
then the value $\theta^{(t)} = \theta'$, and otherwise,
$\theta^{(t)} = \theta^{(t-1)}$.

\section{Multi-fidelity simulated annealing}

In simulated annealing, the goal is to sample from some distribution
\begin{align*}
    P(\theta) \propto \exp(-E(\theta)),
\end{align*}
where $E(\theta)$ is an energy function (the interpretation is, e.g., a negative log-likelihood).
If used for optimization, $E(\theta)$ is the function we are interested in
minimizing.
In the simplest simulated annealing case, we instead sample from the annealed distribution
\begin{align*}
    \pi(\theta) \propto \exp(-E(\theta))^\frac{1}{T}
    = \exp(-E(\theta)/T).
\end{align*}

To adapt this to a multi-fidelity method, we consider
energy functions of fidelity $K$, denoted by $E_K(\theta)$.
the target densities
$\pi(\theta | K)$ and
$\pi(K | \theta)$.

To sample from $\pi(\theta | K)$, we accept or reject a proposal $\theta'$ based on:
\begin{align*}
    R = \exp\left(-\frac{E_K(\theta')-E_K(\theta)}{T} \right).
\end{align*}

To sample from $\pi(K | \theta)$, we accept or reject a proposal $K'$ based on:
\begin{align*}
    R = \exp\left(-\frac{E_{K'}(\theta)-E_{K}(\theta)}{T} \right)
    \left(\frac{\mu(K')}{\mu(K)}\right)^{\frac{1}{T}}.
\end{align*}

\section{Experiments: additional experiments and method details}
\label{appendix:expts}

In this section, we provide additional details for
the methods used in our experiments along with additional
details of the setup of each experiment.

\paragraph{Methods compared}
We will use the abbreviations \emph{SF} to denote a single-fidelity
algorithm, e.g., SF M-H, \emph{MF} to refer to the pseudo-marginal
MF-MCMC method proposed in this work, and \emph{TS} to refer
to the two-stage M-H algorithm described in \Cref{appendix:twostage}.
The primary sampling algorithms used to update the state $\theta | K$ are
Metropolis-Hastings (M-H), (line) slice sampling (SS), and elliptical slice sampling (ESS).

\paragraph{Target estimator $\hat{\pi}$}
In our experiments, by default we consider the Russian roulette estimator
with $\mu = \text{geometric}(\gamma_0)$,
unless stated otherwise.

\paragraph{Sampling the fidelity $K | \theta$}
To sample the fidelity from the conditional target
$K | \theta$,
we consider the following random walk M-H move.
Here the target is
\begin{align}
    \label{eq-condtheta}
    \pi(K | \theta) \propto \mu(K) \hat{\pi}_K(\theta).
\end{align}
To propose a new fidelity, we consider a random walk on the positive
integers: flip a fair coin to determine
a new candidate location $k^* = k \pm 1$, where
$k$ is the current value.
Then we can compute the following ratio and decide to accept/reject
this candidate value:
\begin{align*}
R =  \min\left(1,
    \frac{\mu(k^*) \hat{\pi}_{k^*}(\mathcal{D})} 
    {\mu(k) \hat{\pi}_k(\mathcal{D})}
    \right).
\end{align*}
In problems where the estimator may return negative values,
we compute the absolute value of the estimator $|\hat\pi|$,
as summarized in \Cref{alg:MF-MCMC-sign}.

\subsection{Toy conjugate sequence}

In this example, we consider a toy conjugate Bayesian model,
where the data are assumed to arise i.i.d.\ from
a perfect-fidelity model $L_\infty(\theta) = \mathcal{N}(x; \theta,
\sigma_\infty)$, and
a conjugate prior on $\theta$, $\N(\theta | 0,1)$;
conjugacy leads to a closed form Gaussian posterior density that
we can compute and compare to the posterior samples obtained
from the methods that we compare.
Thus, the perfect-fidelity target is
$\pi_\infty(\theta) \propto \N(\theta | 0,1) \prod_{n=1}^N \N(X_n; \theta,
\sigma_\infty)$.

Now suppose that we only have access to the sequence of
low-fidelity models $L_k(\theta) =  \mathcal{N}(x; \theta, \sigma_k)$,
where $\sigma_k^2 \rightarrow \sigma_\infty^2$.
Here we consider the sequence $\sigma_k^2 = 1 + 2/k^2$
and $\sigma_\infty^2 = 1$.
In this example, we consider the performance of (1)
SF M-H, MF M-H, and two-stage M-H,
and (2)
SF slice sampling and slice sampling (there is not an analogous two-stage MCMC
algorithm for slice sampling).
We generate $N=200$ observations $\mathcal{D}|\theta_0$ from
the perfect-fidelity likelihood with true mean $\theta_0 \sim \N(0,1)$.

To compute  the ``cost'' of a likelihood evaluation,
we pretend that the likelihood evaluation $L_k$ has cost $k$.
This is to demonstrate the cost of the method for problems
where the cost of an evaluation increases linearly with $k$.

In what follows, we first compare the low-fidelity estimators,
and then we compare the sampling methods on one choice of estimator.

\paragraph{Comparing SF-MCMC, MF-MCMC, and two-stage M-H}

We also compare to the two-stage M-H algorithm summarized in \Cref{appendix:twostage}; here we
consider 2 two-stage setups of $k = \{1000,10\}$ and $k = \{100,5\}$.
For all methods, we ran 4 chains initialized from the prior with $T=10000$
iterations.
We discarded 2000 burn-in samples and the subsequently
collected every other sample.

\subsection{Log Gaussian Cox Process}
\label{ssec-appendix-lgcp}

In this section, we provide details for the LGCP experiment on the coal mining disasters data set.

We approximate the integral in \Cref{eq-lgcp} with a trapezoidal quadrature rule
$I_k$: i.e., given $k$ points $\tilde x_1, \ldots, \tilde x_k \in \mathbb{X}$ and observed points
$\{X_1,\ldots,X_N\}$, the low-fidelity likelihood is:
\begin{align}
    \label{eq-lgcp-f}
    L_k(f) = \exp\left(I_k(f(\tilde x_1), \ldots, f(\tilde x_K)) \right)
    \prod_{n=1}^N e^{f(X_n)},
\end{align}
where $I_k$ is a trapezoid quadrature rule with $2k+c$ quadrature points and $c$
is a constant offset parameter.
When computing $L_k$ for a grid of values different than the vector of latent function values
currently available, we draw new function values conditioned on the existing values of $f$.

For all samplers, we used a squared-exponential kernel with lengthscale $\ell =
20$ and variance of 1. For the low-fidelity estimator $\hat{L}_K$, we used a
Russian roulette estimator and set the offset $c=10$.
The truncation parameter of the MF model was fixed at $\gamma_0 = 0.08$.
The results in \Cref{fig:coal} are computed with respect to an average over 4 chains
initialized from the prior with $T=10000$ samples.
The posterior mean estimates were computed after discarding 1000 burnin samples and then collecting every third sample.
The estimates with MF-ESS in \Cref{fig:coal} were adjusted for negative signs;
empirically, we observed roughly $2.5\%$ of negative signs in our experiments.

\subsection{Bayesian ODE system identification}
\label{ssec-appendix-lv-ode}

Given a set of parameters $\theta$  and initial conditions, we can solve the ODE at a fidelity $k$
to obtain the solution $z_n^{(k)}$.
Thus, the likelihood of fidelity $k$ is given by:
\begin{align}
    \label{eq-ode-lowfi}
    L_k(\theta) = \prod_{n=1}^N \prod_{j=1}^2 \text{LogNormal}(\log(z_{n,j}^{(k)}(\theta)), \sigma),
\end{align}
where $k$ represents the fidelity of the ODE solver for obtaining the solution $z_n(\theta)$.
We use the following priors on the parameters
\begin{align}
    (\log\alpha, \log\beta, \log\gamma, \log\delta) \sim \N(\theta_0, \sigma_0 I),
    \qquad \theta_0 = [0, -2, 0, -3]^\top,\quad \sigma_0 = 0.1.
\end{align}

In order to apply elliptical slice sampling, which requires the prior to have mean 0, we apply a
change of variables: define $L_k(\bar{\theta}) = L_k(\theta + \theta_0)$,
and then transform the sampled values $\theta^{(t)} = \bar\theta^{(t)} + \theta_0$.
In our experiments, we first verified the sampler was recovering values on synthetic data
generated
with initial conditions
$z_0 = [1.0,1.0]$, system parameters $\alpha = 1.5,\beta= 1.0, \gamma=3.0,\delta= 1.0$,
and noise parameter $\sigma=0.8$ at a grid of $N$ solution values.

We then applied the method to the Hudson's Bay Lynx-Hare data set, which documents the canadian lynx
and showshoe hare populations between 1900 and 1920, based on the data collefted by the Hudson's Bay
company.
We compared two single-fidelity models with ODE step size $dt = 1\times 10^{-5}, 1 \times 10^{-4}$.
For the multi-fidelity ESS sampler, we visualize the results of $\gamma_0 = 0.12$,
and the step size for the low-fidelity target sequence was computed as
$dt(k) = 1 / (s k + c)$, where we set $s = 10$ and $c = 50$.

\begin{figure}
    \centering
    \begin{subfigure}[b]{\linewidth}
        \centering
        \includegraphics[scale=0.47]{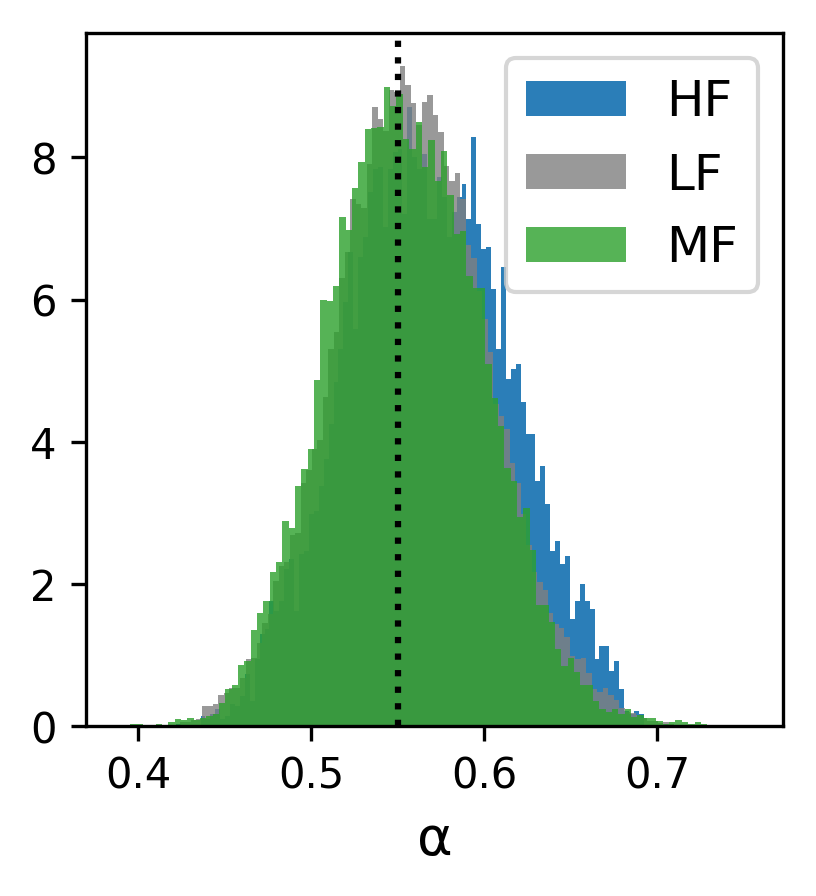}
        \includegraphics[scale=0.47]{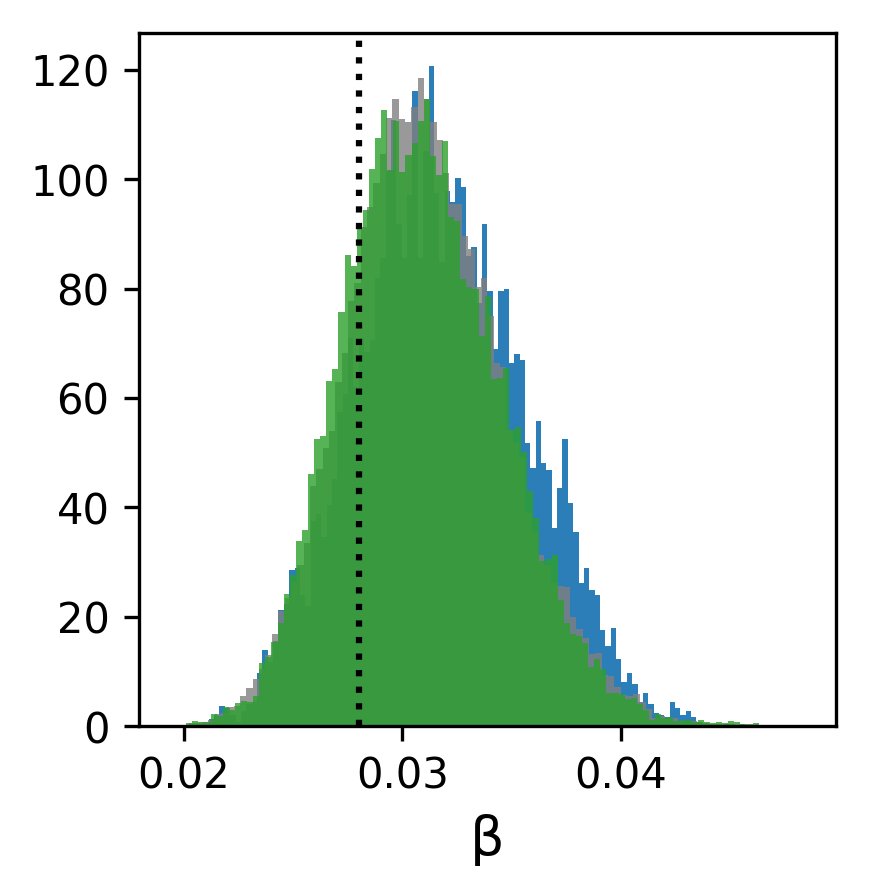}
        \includegraphics[scale=0.47]{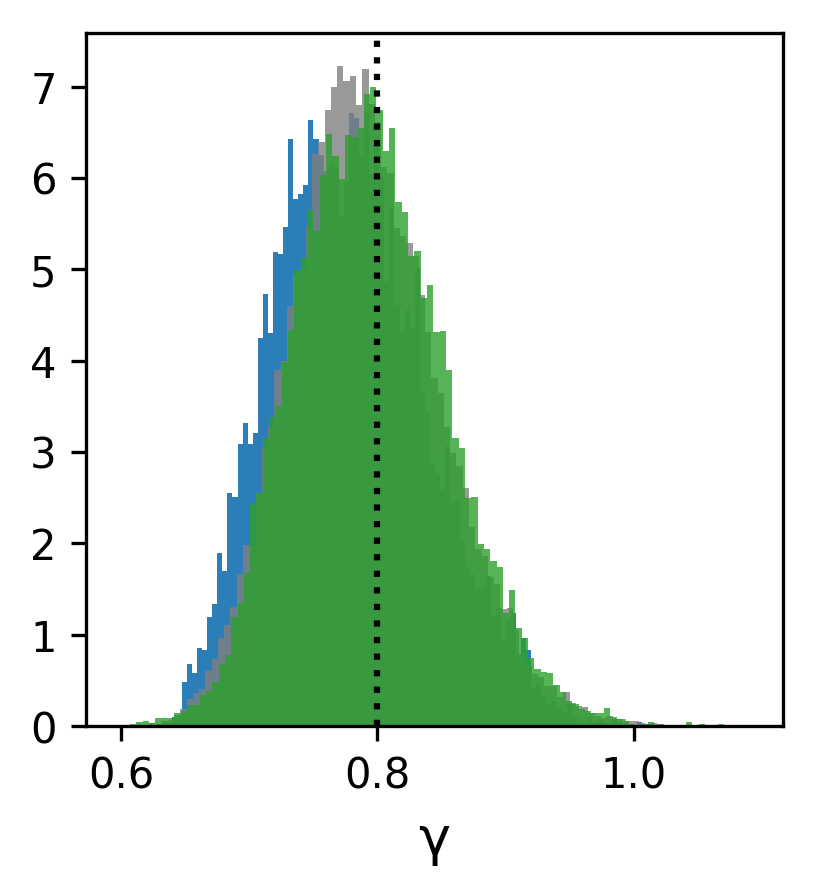}
        \includegraphics[scale=0.47]{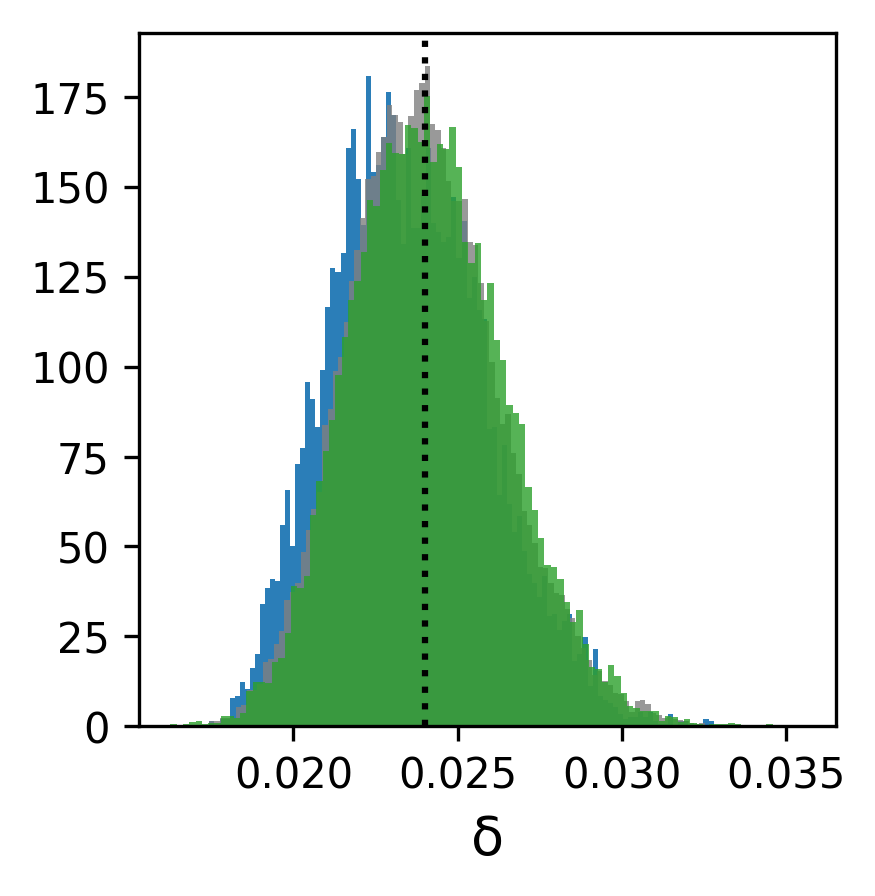}
        \caption{Marginal densities of system parameters}
    \end{subfigure}
    \begin{subfigure}[b]{\linewidth}
       \centering
       \includegraphics[scale=0.65]{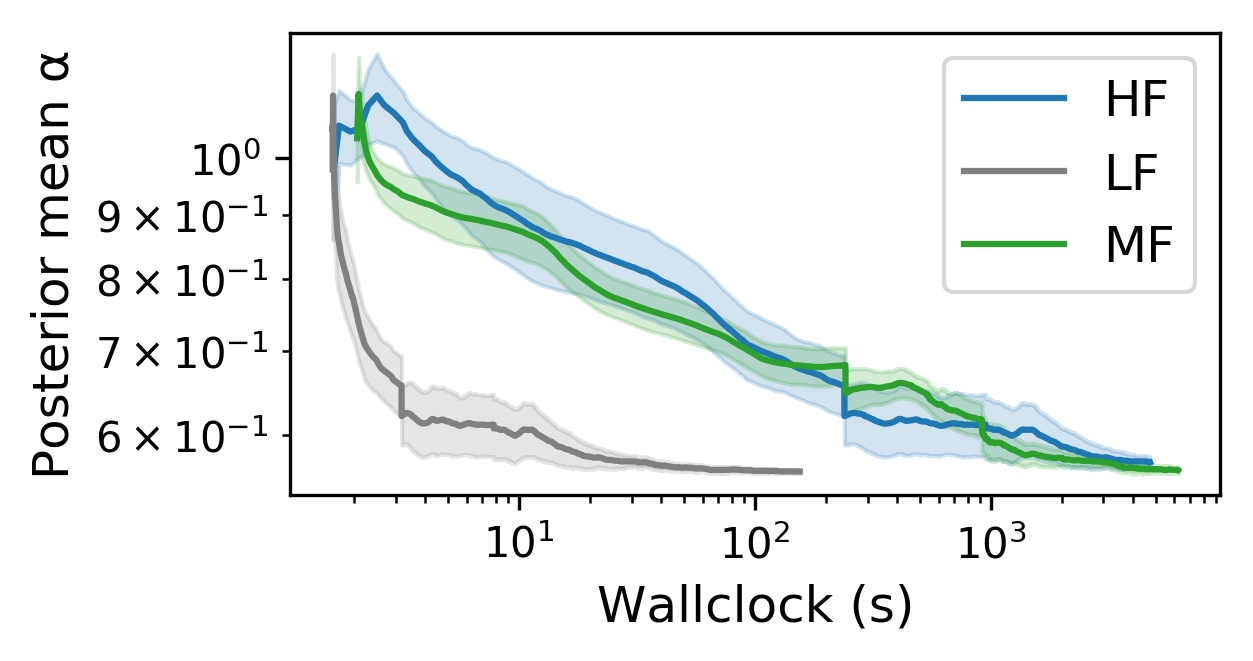}
       \includegraphics[scale=0.65]{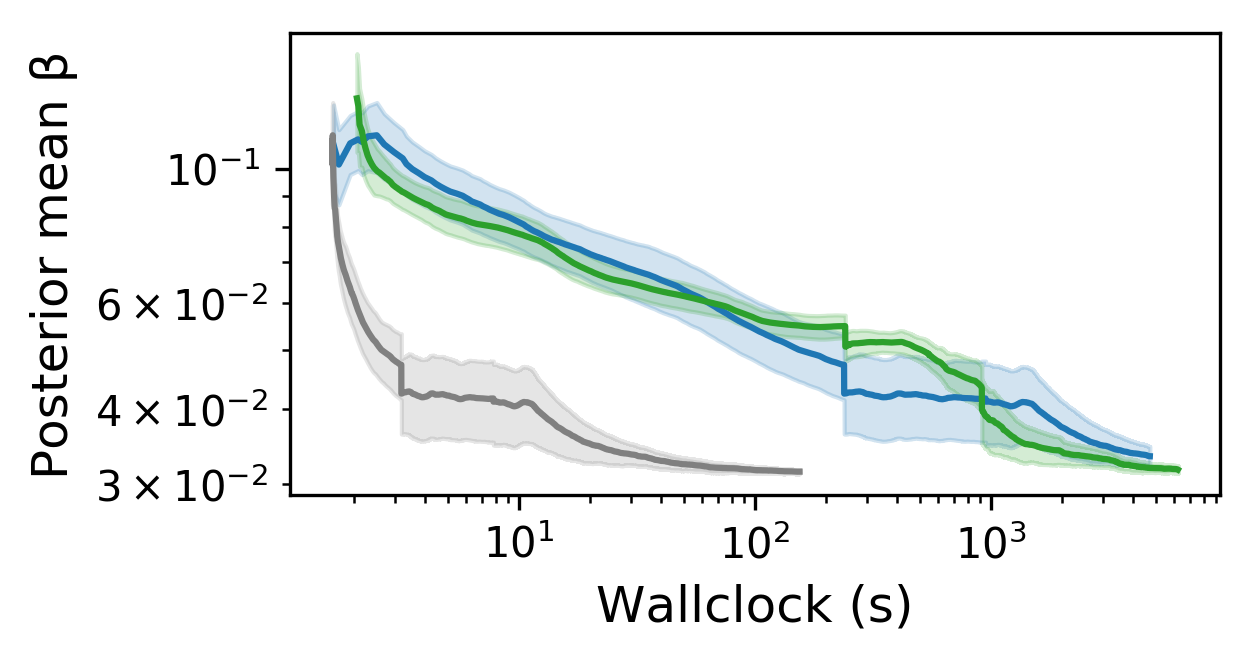}
       \includegraphics[scale=0.65]{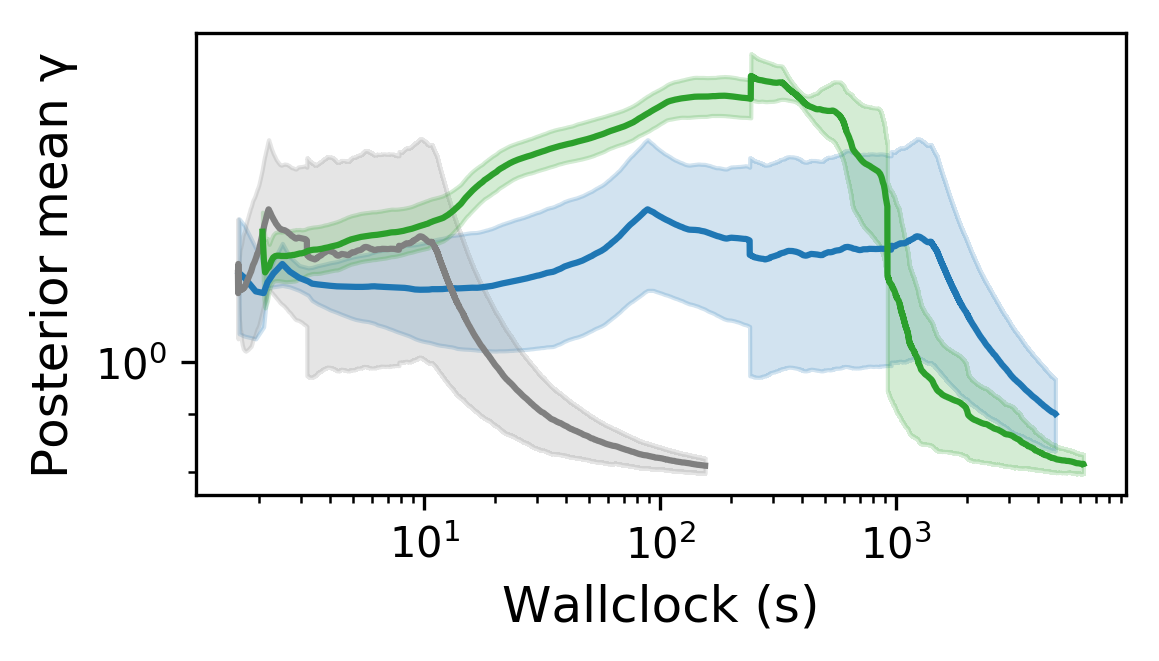}
       \includegraphics[scale=0.65]{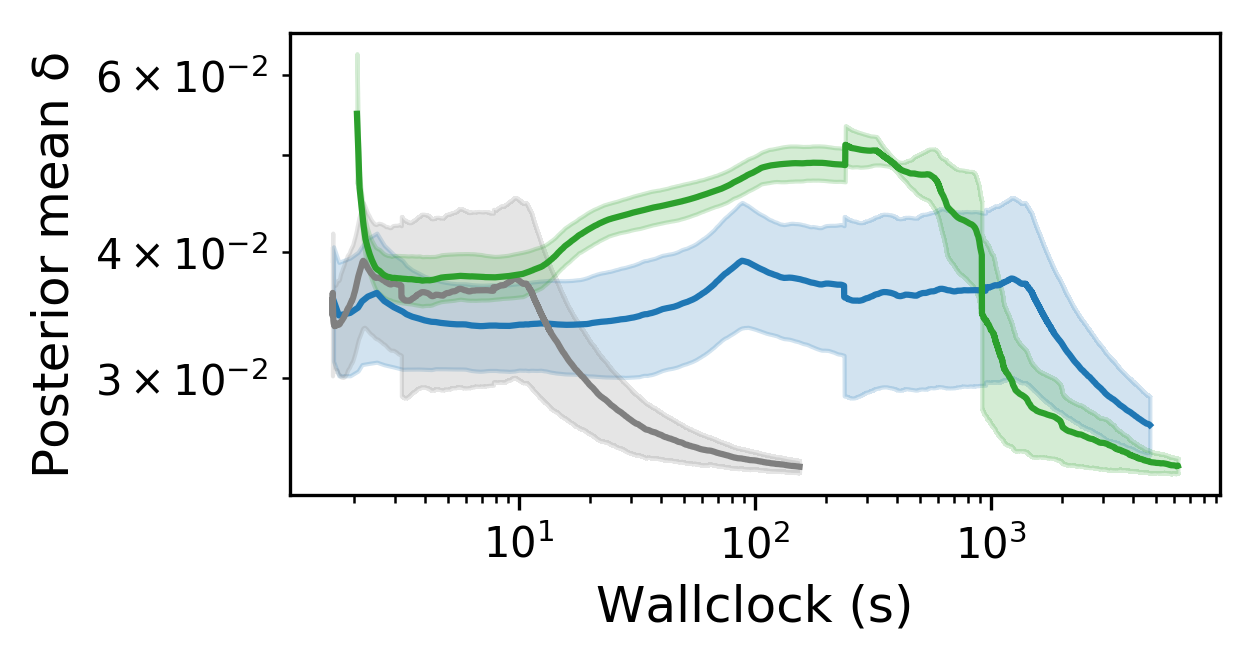}
       \caption{Posterior mean estimate vs computational cost}
    \end{subfigure}
    \caption{Lokta-Volterra system parameter identification
    with a 4th-order Runge Kutta ODE solver.
    The fidelity represents (a function of) the step size of the ODE solver.
    \textbf{Top:} Marginal distributions of system parameters.
    \textbf{Bottom:}
    Posterior mean estimates of the parameters vs wallclock.
    }
    \label{fig:lvode1}
\end{figure}

The results using Euler's method to solve the ODE are in \Cref{fig:lvode},
and the results of the 4th-order Runge Kutta solver are in \Cref{fig:lvode1}.
The maximum number of iterations of each ODE solver was set to $1\times 10^8$ iterations.

In the top row of each figure, the black vertical dotted line denotes maximum likelihood estimates reported by
\citet{howard2009modeling}.%
\footnote{Our model is a modification of the one proposed in a Stan case study, which compares their
Bayesian estimates to the
reported maximum likelihood results. See
\url{https://mc-stan.org/users/documentation/case-studies/lotka-volterra-predator-prey.html} for
further discussion.}
In the bottom row of each figure, we report the posterior mean estimates of the system parameters averaged over
4 chains initialized from the prior.
The wallclock time in seconds of each iteration was measured and the average per iteration was reported.
Here the first 5000 samples of each chain were discarded and then every third sample was collected.
Overall, we observe that the single-fidelity models can both be quite expensive;
while they are able to recover the posterior mean well, they require quite a bit more computation
than the multi-fidelity approach.
Empirically, we observed roughly $1\%$ of negative signs in our experiments.

\subsection{PDE-constrained optimization}

In the problem setting, the spatial domain is $[0,L]$ and the time domain is $[0,T]$.
For our experiments, we chose $L = 10$ and $T=1$.

To solve the PDE, we discretize the spatial domain into a grid of size $\Delta x$: thus, we can consider points
$x_1,\ldots, x_I$ and $u_1(t), \ldots, u_I(t)$, where $u_i(t) = u(x_i, t)$.
Then, we represent the second derivative using the central difference formula for the second degree
derivative:
\begin{align*}
    \frac{\partial^2 u(x,t)}{\partial^2 x} \approx
    \left[\frac{u_{i+1}(t) - 2 u_i(t) + u_{i-1}(t)}{\Delta x^2}\right]_{i=1}^I.
\end{align*}
Thus, we now consider the system of equations (with the appropriate boundary conditions imposed):
\begin{align*}
    \frac{u_{i+1}(t) - 2 u_i(t) + u_{i-1}(t)}{\Delta x^2} = \frac{d u_i(t)}{dt}.
\end{align*}
We solve the system with the Tsitouras 5/4 Runge-Kutta method,
setting $\Delta t = 0.4  \Delta x^2$ so as to satisfy a CFL stability condition.
Here the fidelity of the problem is given by the size of the spatial discretization $\Delta x$,
which in turn controls the discretization of $\Delta t$.

The target temperature $\bar{u}$ was constructed by solving the PDE with parameters $\alpha_0 = 0.85$
and $\beta_0 = 0.21$.
For the simulated annealing algorithm, we use a Metropolis-Hastings algorithm as the base sampler;
all methods used a truncated Normal proposal with scale set to 0.3 and a logarithmic temperature
schedule.

In the top row of
\Cref{fig:pdeconst}, we visualization the target $\bar{u}$ solutions recovered by a number of methods.
The low-fidelity solution in target (c) is given by a crude step size of $\Delta x = 2$; note that
we do not evaluate the cost of this given how poorly the solution is recovered at this state.

In the bottom row of
\Cref{fig:pdeconst}, we compare the MF-ESS approach with two single-fidelity step sizes, $\Delta x =
5 \times 10^{-3}, 1 \times 10^{-2}$.
In the multi-fidelity method, the low-fidelity target sequence was chosen using the discretization
sequence $\Delta x(k) = 1 / (k + c)$, where $c = 8$.
The results are averaged over random seeds using the initialization $[0,0]$.
The horizontal dotted lines in each plot  denote the values of $\alpha_0, \beta_0$,
and we plot the current minimum at each iteration.

\subsection{Gaussian process regression parameter inference}
\label{ssec-appendix-gp}

In many applications of GPs, the goal is to integrate out the parameters $\theta$ via
a Monte Carlo approximation that uses MCMC to sample $\{\theta^{(t)}\}$ from
the target density
\begin{align}
    \label{eq:posterior}
    \pi_\infty(\theta \given \mathcal{D} = (X,y)) \propto \pi(\theta) L_\infty(\theta)
    = \text{logNormal}(\theta \given \nu_0,\nu_1) \times \N(y \given 0, \Sigma_{\theta} + \sigma_0^2 I).
\end{align}
Note that the Gaussian pdf has the form
\begin{align}
    L_\infty(\theta)  = |2\pi(\Sigma_{\theta} + \sigma_0^2 I)|^{-\frac{1}{2}}
    \exp\left(-\frac{1}{2} y^\top (\Sigma_{\theta} + \sigma_0^2 I)^{-1} y\right),
\end{align}
and so when $N$ is large, the linear system and determinant above become expensive.

Let the low-fidelity likelihood $L_k(\theta)$ denote the computation of
the likelihood with $k$ iterations of (preconditioned) conjugate gradient.
That is, suppose, $z^{(k)}$ is the $k^{th}$ iteration of the CG with respect to the linear system
$(\Sigma_{\theta} + \sigma_0^2 I)z = y$.
Thus, the low-fidelity likelihood is
\begin{align*}
    L_k(\theta) = |2\pi(\Sigma_{\theta} + \sigma_0^2 I)|^{-\frac{1}{2}}
    \exp\left(-\frac{1}{2} y^\top z^{(k)}\right).
\end{align*}
In practice, the determinant also needs to be approximated
with another low-fidelity computation.
Our goal here is to show a proof of concept, and so
we only consider the linear system above; however,
we note that the determinant can be iteratively computed as
a byproduct of conjugate gradient as in \citet{potapczynski2021bias}.
Note that we can compute the likelihood recursively in that each $z^{(k)}$ reuses computation from
the previous step $z^{(k-1)}$, and thus a Russian roulette estimator also can reuse computation for
each term in the sum.

We generate synthetic data from the GP model with $N=100$,
$\sigma_0^2 = 1$, and lengthscale $\theta_0 = 45$.
For the GP model, we use the Log Normal prior on $\theta$ given above in
\Cref{eq:posterior} with parameters $\nu_0 = 3.8, \nu_1 = 0.03$.
We compare several likelihoods:
a high-fidelity likelihood ($K=100$),
low-fidelity likelihood ($K = 5$),
and
the multi-fidelity approach we describe with $\gamma_0 = 0.1$.
The low-fidelity likelihood sequence was constructed by computing the solution to the linear system
using a conjugate gradient solver with $k$ steps.
Finally, we also compare to a two-stage M-H approach with $k \in \{100,5\}$.
For all methods, we use a M-H sampler with $T=50000$ iterations.
The results are in \Cref{fig:GPreg}.

\bibliographystyle{plainnat-mod}
\bibliography{main}

\begin{thebibliography}{52}
\providecommand{\natexlab}[1]{#1}
\providecommand{\url}[1]{\texttt{#1}}
\expandafter\ifx\csname urlstyle\endcsname\relax
  \providecommand{\doi}[1]{doi: #1}\else
  \providecommand{\doi}{doi: \begingroup \urlstyle{rm}\Url}\fi

\bibitem[Alexandrov et~al.(1998)Alexandrov, Dennis, Lewis, and
  Torczon]{alexandrov1998trust}
N.~M. Alexandrov, J.~E. Dennis, R.~M. Lewis, and V.~Torczon.
\newblock A trust-region framework for managing the use of approximation models
  in optimization.
\newblock \emph{Structural optimization}, 15\penalty0 (1):\penalty0 16--23,
  1998.

\bibitem[Andrieu and Roberts(2009)]{andrieu2009pseudo}
C.~Andrieu and G.~O. Roberts.
\newblock The pseudo-marginal approach for efficient {M}onte {C}arlo
  computations.
\newblock \emph{The Annals of Statistics}, 37\penalty0 (2):\penalty0 697--725,
  2009.

\bibitem[Andrieu et~al.(2010)Andrieu, Doucet, and Holenstein]{leeholmes2010}
C.~Andrieu, A.~Doucet, and R.~Holenstein.
\newblock Discussion of ``{P}article {M}arkov chain {M}onte {C}arlo methods".
\newblock \emph{Journal of the Royal Statistical Society: Series B (Statistical
  Methodology)}, 72\penalty0 (3):\penalty0 269--342, 2010.

\bibitem[Arian et~al.(2000)Arian, Fahl, and Sachs]{arian2000trust}
E.~Arian, M.~Fahl, and E.~W. Sachs.
\newblock Trust-region proper orthogonal decomposition for flow control.
\newblock In \emph{IEEE Conference on Decision and Control}, 2000.

\bibitem[Beatson and Adams(2019)]{beatson2019efficient}
A.~Beatson and R.~P. Adams.
\newblock Efficient optimization of loops and limits with randomized
  telescoping sums.
\newblock In \emph{International Conference on Machine Learning}, pages
  534--543. PMLR, 2019.

\bibitem[Beaumont(2003)]{beaumont2003estimation}
M.~A. Beaumont.
\newblock Estimation of population growth or decline in genetically monitored
  populations.
\newblock \emph{Genetics}, 164\penalty0 (3):\penalty0 1139--1160, 2003.

\bibitem[Biron-Lattes et~al.(2022)Biron-Lattes, Bouchard-C{\^o}t{\'e}, and
  Campbell]{biron2021pseudo}
M.~Biron-Lattes, A.~Bouchard-C{\^o}t{\'e}, and T.~Campbell.
\newblock Pseudo-marginal inference for {CTMC}s on infinite spaces via
  monotonic likelihood approximations.
\newblock \emph{Journal of Computational and Graphical Statistics}, 2022.

\bibitem[Brevault et~al.(2020)Brevault, Balesdent, and
  Hebbal]{brevault2020overview}
L.~Brevault, M.~Balesdent, and A.~Hebbal.
\newblock Overview of {G}aussian process based multi-fidelity techniques with
  variable relationship between fidelities.
\newblock \emph{arXiv preprint arXiv:2006.16728}, 2020.

\bibitem[Carlin et~al.(1992)Carlin, Gelfand, and Smith]{carlin1992hierarchical}
B.~P. Carlin, A.~E. Gelfand, and A.~F. Smith.
\newblock Hierarchical {B}ayesian analysis of changepoint problems.
\newblock \emph{Journal of the royal statistical society: series C (applied
  statistics)}, 41\penalty0 (2):\penalty0 389--405, 1992.

\bibitem[Christen and Fox(2005)]{christen2005markov}
J.~A. Christen and C.~Fox.
\newblock Markov chain {M}onte {C}arlo using an approximation.
\newblock \emph{Journal of Computational and Graphical Statistics}, 14\penalty0
  (4):\penalty0 795--810, 2005.

\bibitem[Cui et~al.(2015)Cui, Marzouk, and Willcox]{cui2015data}
T.~Cui, Y.~M. Marzouk, and K.~E. Willcox.
\newblock Data-driven model reduction for the {B}ayesian solution of inverse
  problems.
\newblock \emph{International Journal for Numerical Methods in Engineering},
  102\penalty0 (5):\penalty0 966--990, 2015.

\bibitem[Diggle et~al.(2013)Diggle, Moraga, Rowlingson, and
  Taylor]{diggle2013spatial}
P.~J. Diggle, P.~Moraga, B.~Rowlingson, and B.~M. Taylor.
\newblock Spatial and spatio-temporal log-{G}aussian {C}ox processes: extending
  the geostatistical paradigm.
\newblock \emph{Statistical Science}, 28\penalty0 (4):\penalty0 542--563, 2013.

\bibitem[Dodwell et~al.(2015)Dodwell, Ketelsen, Scheichl, and
  Teckentrup]{dodwell2015hierarchical}
T.~J. Dodwell, C.~Ketelsen, R.~Scheichl, and A.~L. Teckentrup.
\newblock A hierarchical multilevel {Markov chain Monte Carlo} algorithm with
  applications to uncertainty quantification in subsurface flow.
\newblock \emph{SIAM/ASA Journal on Uncertainty Quantification}, 3\penalty0
  (1):\penalty0 1075--1108, 2015.

\bibitem[Efendiev et~al.(2006)Efendiev, Hou, and
  Luo]{efendiev2006preconditioning}
Y.~Efendiev, T.~Hou, and W.~Luo.
\newblock Preconditioning {M}arkov chain {M}onte {C}arlo simulations using
  coarse-scale models.
\newblock \emph{SIAM Journal on Scientific Computing}, 28\penalty0
  (2):\penalty0 776--803, 2006.

\bibitem[Fahl and Sachs(2003)]{fahl2003reduced}
M.~Fahl and E.~W. Sachs.
\newblock Reduced order modelling approaches to {PDE}-constrained optimization
  based on proper orthogonal decomposition.
\newblock In \emph{Large-scale PDE-constrained optimization}, pages 268--280.
  Springer, 2003.

\bibitem[Georgoulas et~al.(2017)Georgoulas, Hillston, and
  Sanguinetti]{georgoulas2017unbiased}
A.~Georgoulas, J.~Hillston, and G.~Sanguinetti.
\newblock Unbiased {B}ayesian inference for population {M}arkov jump processes
  via random truncations.
\newblock \emph{Statistics and computing}, 27\penalty0 (4):\penalty0 991--1002,
  2017.

\bibitem[Gessner et~al.(2020)Gessner, Gonzalez, and
  Mahsereci]{gessner2020active}
A.~Gessner, J.~Gonzalez, and M.~Mahsereci.
\newblock Active multi-information source {B}ayesian quadrature.
\newblock In \emph{Uncertainty in Artificial Intelligence}, pages 712--721.
  PMLR, 2020.

\bibitem[Giles(2008)]{giles2008multilevel}
M.~B. Giles.
\newblock Multilevel {M}onte {C}arlo path simulation.
\newblock \emph{Operations research}, 56\penalty0 (3):\penalty0 607--617, 2008.

\bibitem[Giles(2013)]{giles2013multilevel}
M.~B. Giles.
\newblock Multilevel {M}onte {C}arlo methods.
\newblock \emph{Monte Carlo and Quasi-Monte Carlo Methods}, pages 83--103,
  2013.

\bibitem[Glynn and Rhee(2014)]{glynn2014exact}
P.~W. Glynn and C.-h. Rhee.
\newblock Exact estimation for {M}arkov chain equilibrium expectations.
\newblock \emph{Journal of Applied Probability}, 51\penalty0 (A):\penalty0
  377--389, 2014.

\bibitem[Gramacy and Lee(2009)]{gramacy2009adaptive}
R.~B. Gramacy and H.~K. Lee.
\newblock Adaptive design and analysis of supercomputer experiments.
\newblock \emph{Technometrics}, 51\penalty0 (2):\penalty0 130--145, 2009.

\bibitem[Guha and Tan(2017)]{guha2017multilevel}
N.~Guha and X.~Tan.
\newblock Multilevel approximate {B}ayesian approaches for flows in highly
  heterogeneous porous media and their applications.
\newblock \emph{Journal of Computational and Applied Mathematics},
  317:\penalty0 700--717, 2017.

\bibitem[Gundersen et~al.(2021)Gundersen, Cai, Zhou, Engelhardt, and
  Adams]{gundersen2021active}
G.~W. Gundersen, D.~Cai, C.~Zhou, B.~E. Engelhardt, and R.~P. Adams.
\newblock Active multi-fidelity {B}ayesian online changepoint detection.
\newblock In \emph{Uncertainty in Artificial Intelligence}, pages 1916--1926.
  PMLR, 2021.

\bibitem[Hackbusch(2013)]{hackbusch2013multi}
W.~Hackbusch.
\newblock \emph{Multi-grid methods and applications}, volume~4.
\newblock Springer Science \& Business Media, 2013.

\bibitem[Higdon et~al.(2002)Higdon, Lee, and Bi]{higdon2002bayesian}
D.~Higdon, H.~Lee, and Z.~Bi.
\newblock A {B}ayesian approach to characterizing uncertainty in inverse
  problems using coarse and fine-scale information.
\newblock \emph{IEEE Transactions on Signal Processing}, 50\penalty0
  (2):\penalty0 389--399, 2002.

\bibitem[Howard(2009)]{howard2009modeling}
P.~Howard.
\newblock Modeling basics.
\newblock \emph{Lecture Notes for Math}, 442, 2009.

\bibitem[Jacob et~al.(2020)Jacob, O’Leary, and
  Atchad{\'e}]{jacob2020unbiased}
P.~E. Jacob, J.~O’Leary, and Y.~F. Atchad{\'e}.
\newblock Unbiased {M}arkov chain {M}onte {C}arlo methods with couplings.
\newblock \emph{Journal of the Royal Statistical Society: Series B (Statistical
  Methodology)}, 82\penalty0 (3):\penalty0 543--600, 2020.

\bibitem[Jones et~al.(1998)Jones, Schonlau, and Welch]{jones1998efficient}
D.~R. Jones, M.~Schonlau, and W.~J. Welch.
\newblock Efficient global optimization of expensive black-box functions.
\newblock \emph{Journal of Global optimization}, 13\penalty0 (4):\penalty0
  455--492, 1998.

\bibitem[Lester(2018)]{lester2018multi}
C.~Lester.
\newblock Multi-level approximate {B}ayesian computation.
\newblock \emph{arXiv preprint arXiv:1811.08866}, 2018.

\bibitem[Li et~al.(2020)Li, Xing, Kirby, and Zhe]{li2020multi}
S.~Li, W.~Xing, R.~Kirby, and S.~Zhe.
\newblock Multi-fidelity {B}ayesian optimization via deep neural networks.
\newblock \emph{Advances in Neural Information Processing Systems},
  33:\penalty0 8521--8531, 2020.

\bibitem[Lin et~al.(2000)Lin, Liu, and Sloan]{lin2000noisy}
L.~Lin, K.~Liu, and J.~Sloan.
\newblock A noisy {M}onte {C}arlo algorithm.
\newblock \emph{Physical Review D}, 61\penalty0 (7):\penalty0 074505, 2000.

\bibitem[Luo et~al.(2020)Luo, Beatson, Norouzi, Zhu, Duvenaud, Adams, and
  Chen]{luo2020sumo}
Y.~Luo, A.~Beatson, M.~Norouzi, J.~Zhu, D.~Duvenaud, R.~P. Adams, and R.~T.
  Chen.
\newblock {SUMO}: {U}nbiased estimation of log marginal probability for latent
  variable models.
\newblock \emph{ICLR}, 2020.

\bibitem[Lyne et~al.(2015)Lyne, Girolami, Atchad{\'e}, Strathmann, and
  Simpson]{lyne2015russian}
A.-M. Lyne, M.~Girolami, Y.~Atchad{\'e}, H.~Strathmann, and D.~Simpson.
\newblock On {R}ussian roulette estimates for {B}ayesian inference with
  doubly-intractable likelihoods.
\newblock \emph{Statistical Science}, 30\penalty0 (4):\penalty0 443--467, 2015.

\bibitem[MacKay et~al.(2003)MacKay, Mac~Kay, et~al.]{mackay2003information}
D.~J. MacKay, D.~J. Mac~Kay, et~al.
\newblock \emph{Information theory, inference and learning algorithms}.
\newblock Cambridge University Press, 2003.

\bibitem[March and Willcox(2012)]{march2012constrained}
A.~March and K.~Willcox.
\newblock Constrained multifidelity optimization using model calibration.
\newblock \emph{Structural and Multidisciplinary Optimization}, 46\penalty0
  (1):\penalty0 93--109, 2012.

\bibitem[M{\o}ller et~al.(1998)M{\o}ller, Syversveen, and
  Waagepetersen]{moller1998log}
J.~M{\o}ller, A.~R. Syversveen, and R.~P. Waagepetersen.
\newblock Log {G}aussian {C}ox processes.
\newblock \emph{Scandinavian Journal of Statistics}, 25\penalty0 (3):\penalty0
  451--482, 1998.

\bibitem[Murray and Graham(2016)]{murray2016pseudo}
I.~Murray and M.~Graham.
\newblock Pseudo-marginal slice sampling.
\newblock In \emph{Artificial Intelligence and Statistics}, pages 911--919.
  PMLR, 2016.

\bibitem[Murray et~al.(2010)Murray, Adams, and MacKay]{murray2010elliptical}
I.~Murray, R.~Adams, and D.~MacKay.
\newblock Elliptical slice sampling.
\newblock In \emph{Artificial Intelligence and Statistics}, pages 541--548.
  PMLR, 2010.

\bibitem[Neal(2003)]{neal2003slice}
R.~M. Neal.
\newblock Slice sampling.
\newblock \emph{The Annals of Statistics}, 31\penalty0 (3):\penalty0 705--767,
  2003.

\bibitem[Palizhati et~al.(2022)Palizhati, Torrisi, Aykol, Suram, Hummelsh{\o}j,
  and Montoya]{palizhati2022agents}
A.~Palizhati, S.~B. Torrisi, M.~Aykol, S.~K. Suram, J.~S. Hummelsh{\o}j, and
  J.~H. Montoya.
\newblock Agents for sequential learning using multiple-fidelity data.
\newblock \emph{Scientific reports}, 12\penalty0 (1):\penalty0 1--13, 2022.

\bibitem[Peherstorfer et~al.(2018)Peherstorfer, Willcox, and
  Gunzburger]{peherstorfer2018survey}
B.~Peherstorfer, K.~Willcox, and M.~Gunzburger.
\newblock Survey of multifidelity methods in uncertainty propagation,
  inference, and optimization.
\newblock \emph{SIAM Review}, 60\penalty0 (3):\penalty0 550--591, 2018.

\bibitem[Potapczynski et~al.(2021)Potapczynski, Wu, Biderman, Pleiss, and
  Cunningham]{potapczynski2021bias}
A.~Potapczynski, L.~Wu, D.~Biderman, G.~Pleiss, and J.~P. Cunningham.
\newblock Bias-free scalable {G}aussian processes via randomized truncations.
\newblock In \emph{International Conference on Machine Learning}, pages
  8609--8619. PMLR, 2021.

\bibitem[Prescott and Baker(2020)]{prescott2020multifidelity}
T.~P. Prescott and R.~E. Baker.
\newblock Multifidelity approximate {B}ayesian computation.
\newblock \emph{SIAM/ASA Journal on Uncertainty Quantification}, 8\penalty0
  (1):\penalty0 114--138, 2020.

\bibitem[Raissi et~al.(2017)Raissi, Perdikaris, and
  Karniadakis]{raissi2017inferring}
M.~Raissi, P.~Perdikaris, and G.~E. Karniadakis.
\newblock Inferring solutions of differential equations using noisy
  multi-fidelity data.
\newblock \emph{Journal of Computational Physics}, 335:\penalty0 736--746,
  2017.

\bibitem[Robinson et~al.(2008)Robinson, Eldred, Willcox, and
  Haimes]{robinson2008surrogate}
T.~Robinson, M.~S. Eldred, K.~E. Willcox, and R.~Haimes.
\newblock Surrogate-based optimization using multifidelity models with variable
  parameterization and corrected space mapping.
\newblock \emph{AIAA journal}, 46\penalty0 (11):\penalty0 2814--2822, 2008.

\bibitem[Romeijn and Smith(1994)]{romeijn1994simulated}
H.~E. Romeijn and R.~L. Smith.
\newblock Simulated annealing for constrained global optimization.
\newblock \emph{Journal of Global Optimization}, 5\penalty0 (2):\penalty0
  101--126, 1994.

\bibitem[Song et~al.(2019)Song, Chen, and Yue]{song2019general}
J.~Song, Y.~Chen, and Y.~Yue.
\newblock A general framework for multi-fidelity {B}ayesian optimization with
  gaussian processes.
\newblock In \emph{The 22nd International Conference on Artificial Intelligence
  and Statistics}, pages 3158--3167. PMLR, 2019.

\bibitem[Teng et~al.(2017)Teng, Nathoo, and Johnson]{teng2017bayesian}
M.~Teng, F.~Nathoo, and T.~D. Johnson.
\newblock Bayesian computation for {L}og-{G}aussian {C}ox processes: A
  comparative analysis of methods.
\newblock \emph{Journal of Statistical Computation and Simulation}, 87\penalty0
  (11):\penalty0 2227--2252, 2017.

\bibitem[Troyer and Wiese(2005)]{troyer2005computational}
M.~Troyer and U.-J. Wiese.
\newblock Computational complexity and fundamental limitations to fermionic
  quantum {M}onte {C}arlo simulations.
\newblock \emph{Physical Review Letters}, 94\penalty0 (17):\penalty0 170201,
  2005.

\bibitem[Warne et~al.(2021)Warne, Prescott, Baker, and
  Simpson]{warne2021multifidelity}
D.~J. Warne, T.~P. Prescott, R.~E. Baker, and M.~J. Simpson.
\newblock Multifidelity multilevel {M}onte {C}arlo to accelerate approximate
  {B}ayesian parameter inference for partially observed stochastic processes.
\newblock \emph{arXiv preprint arXiv:2110.14082}, 2021.

\bibitem[Wu et~al.(2020)Wu, Toscano-Palmerin, Frazier, and
  Wilson]{wu2020practical}
J.~Wu, S.~Toscano-Palmerin, P.~I. Frazier, and A.~G. Wilson.
\newblock Practical multi-fidelity {B}ayesian optimization for hyperparameter
  tuning.
\newblock In \emph{Uncertainty in Artificial Intelligence}, pages 788--798.
  PMLR, 2020.

\bibitem[Xi et~al.(2018)Xi, Briol, and Girolami]{xi2018bayesian}
X.~Xi, F.-X. Briol, and M.~Girolami.
\newblock Bayesian quadrature for multiple related integrals.
\newblock In \emph{International Conference on Machine Learning}, pages
  5373--5382. PMLR, 2018.

\end{thebibliography}

\end{document}